\documentclass[lettersize,journal]{IEEEtran}
\usepackage{amsmath,amsfonts,amssymb,amsthm}
\usepackage{algorithmic}
\usepackage{algorithm}
\usepackage{array}
\usepackage[caption=false,labelfont=sf,textfont=sf]{subfig}
\usepackage{textcomp}
\usepackage{stfloats}
\usepackage{url}
\usepackage{verbatim}
\usepackage{graphicx}
\usepackage{cite}
\usepackage[dvipsnames]{xcolor}
\usepackage{lipsum}
\usepackage{enumerate}
\usepackage{enumitem}
\usepackage{booktabs}

\DeclareMathOperator*{\argmax}{arg\,max}
\DeclareMathOperator*{\argmin}{arg\,min}

\newtheorem{definition}{Definition}

\begin{document}

\title{GUIDE-VAE: Advancing Data Generation with User Information and Pattern Dictionaries}

\author{Kutay Bölat, Simon~H.~Tindemans
\thanks{The authors are with the Department of Electrical Sustainable Energy, Delft University of Technology, Mekelweg 4, 2628 CD Delft, The Netherlands.}%
\thanks{This research was undertaken as part of the InnoCyPES project, which has received funding from the European Union's Horizon 2020 research and innovation programme under the Marie Sk{\l}odowska-Curie grant agreement No 956433.}}

\markboth{Journal of \LaTeX\ Class Files,~Vol.~14, No.~8, August~2021}%
{Shell \MakeLowercase{\textit{et al.}}: A Sample Article Using IEEEtran.cls for IEEE Journals}

\maketitle

\begin{abstract}
Generative modelling of multi-user datasets has become prominent in science and engineering. Generating a data point for a given user requires employing user information, and conventional generative models, including variational autoencoders (VAEs), often ignore this. This paper introduces GUIDE-VAE, a novel conditional generative model that leverages user embeddings to generate user-guided data. By leveraging shared patterns across users, GUIDE-VAE improves performance in multi-user settings, even under significant data imbalance. In addition to integrating user information, GUIDE-VAE incorporates a pattern dictionary-based covariance composition (PDCC) to improve the realism of generated samples by capturing complex feature dependencies. While user embeddings drive performance gains, PDCC addresses common issues such as noise and over-smoothing typically seen in VAEs.

The proposed GUIDE-VAE was evaluated on a multi-user smart meter dataset characterised by substantial data imbalance across users. Quantitative results show that GUIDE-VAE performs effectively on both synthetic data generation and missing-record imputation tasks, while qualitative evaluations indicate that it produces more plausible and less noisy data. These results establish GUIDE-VAE as a promising tool for controlled, realistic data generation in multi-user datasets, with potential applications across domains that require user-informed modelling.
\end{abstract}

\begin{IEEEkeywords}
Covariance structure, generative modelling, multi-user datasets, user embedding, variational autoencoders
\end{IEEEkeywords}

\section{Introduction}

The landscape of generative artificial intelligence grows at an ever-increasing pace. The underlying generative models continue to expand in size and capacity each year, enabling the generation of novel content (text, images, audio, etc.) with performance close to, if not better than, human-level. Beyond content creation, generative models are now being leveraged by scientists, engineers, and businesses for applications such as synthetic data generation \cite{qian2023synthetic} and scientific discovery \cite{peng2022pocket2mol}. These applications often require some control over the generation mechanism to preserve the utility of the generated samples. Conditional generative models offer such control by introducing desired conditions to the generative process. While contemporary large language models incorporate text prompts for this purpose, datasets from various scientific domains (such as measurements, logs, and surveys) may require more tailored conditions \cite{yoon2020anonymization}.

Variational autoencoders (VAEs) \cite{kingma2013auto}, along with conditional VAEs (CVAEs) \cite{sohn2015learning}, are widely adopted generative models that utilise probabilistic encoders and decoders. These models extend variational inference methods through the power of deep neural networks, offering notable advantages, e.g. efficient data encoding and a strong probabilistic foundation. However, similar to many generative models, VAEs have certain limitations. One well-known drawback is that they are prone to producing synthetic samples that lack fine-grained detail, often resulting in blurry or noisy outputs \cite{bredell2023explicitly}. This issue is particularly striking in time-series data, where strong feature correlations are common, and failing to capture them can significantly reduce the realism of generated samples.

\begin{figure}[t!]
    \centering
    \includegraphics[trim={1.5cm 0 0 0},clip,width=\linewidth]{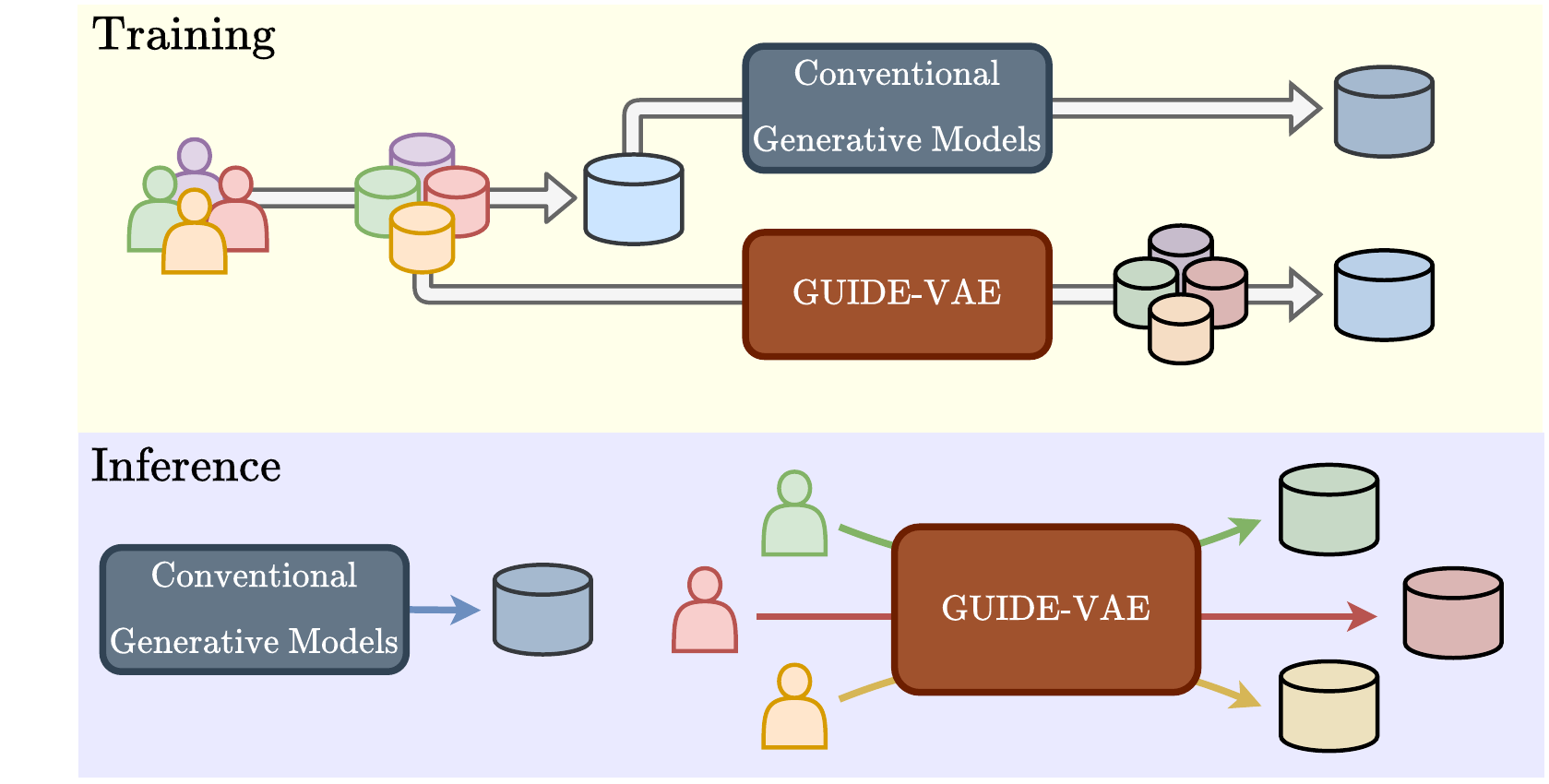}
    \caption{Conventional generative models disregard user information during training, treating the dataset anonymously, which limits their ability to generate data for specific users during inference. GUIDE-VAE addresses this by incorporating user information in the training process, enabling control over the generated outputs for individual users.}
    \label{fig:idea_diagram}
\end{figure}

Multi-user datasets, such as smart meter readings \cite{chai2024defining}, patient health records \cite{sun2023generating}, and user interaction data from digital platforms \cite{zhang2019deep}, consist of records from multiple individuals, with each user contributing data that reflects their unique behaviour and preferences. These datasets are crucial for capturing user-level patterns, such as individual energy consumption \cite{chen2022constructing} and patient treatment responses \cite{sun2023generating}, which are informative for personalised recommendations and targeted solutions. In this context, synthetic data generation enables the analysis and simulation of these trends \cite{almansoori2022padpaf}. However, modelling these datasets presents significant challenges: training separate models for each user is inefficient and impractical, while anonymising user identities leads to a loss of control over generating the specific patterns unique to each user, as shown in Fig.~\ref{fig:idea_diagram}.

One well-known data mining application parallel to multi-user datasets is topic modelling of textual data. This technique analyses a corpus of documents, each consisting of multiple words, and models their relationships by assigning abstract topics to documents. Latent Dirichlet allocation (LDA) \cite{blei2003latent} is a widely used method for topic modelling that effectively captures the latent structure of the data. The resemblance between a document corpus and a multi-user dataset—where documents correspond to users and words to individual data records—makes the application of LDA in multi-user domains both intuitive and powerful. This analogy enables LDA to model the structure of multi-user datasets by grouping similar users based on their data characteristics. In \cite{chen2022constructing}, this approach is applied to smart meter data, yielding useful insights into consumer behaviour and consumption patterns. LDA's ability to reveal such underlying patterns makes it a promising tool for extracting valuable information from multi-user datasets across diverse applications.

\subsection{Problem Statement}
This work addresses two main challenges: (i) the absence of user-specific guidance during the generative process and (ii) the lack of realism in the generated data. The following subsections delve deeper into these issues.

\subsubsection{Lack of Guidance}
Efficient application of conventional generative models to multi-user datasets remains limited. These datasets contain user-specific information, which, if overlooked, can lead to synthetic data that fails to accurately reflect individual patterns and behaviours. This problem becomes even more severe in the presence of data imbalance, where users with lower data counts are underrepresented \cite{sun2023generating} and may hinder fairness \cite{van2021decaf}. Moreover, without incorporating user information, controlling the generative process to produce data for a specific user becomes highly challenging and may require complex, ad-hoc methods \cite{hans2024tabular}.

The key issue is the lack of a mechanism to represent user identities as mathematical objects that can serve as conditions during the generative process. Without such a representation, generative models cannot efficiently encode or utilise the user-specific information required to guide data generation. A similar problem arises in the neural collaborative filtering field, where users and the items they interact with are related to each other \cite{zhang2019deep}. This requires representing users as inputs to neural networks. However, this is generally done by using one-hot encoding, which limits the scalability and generalisation.

To address this challenge, user embeddings that represent users based on their data offer a promising solution. These embeddings map users to a metric space where their similarities can be quantified \cite{chen2022constructing}, allowing the generative model to condition on them and generate data that reflects user-level patterns. However, conventional generative models do not employ such an approach, leaving them unable to accurately condition on user identities. The absence of this capability not only limits the precision of the generated data but also hinders the potential for inter-user knowledge transfer, thereby reducing the quality and variability of the synthetic outputs.

\subsubsection{Lack of Realism}\label{sec:lack_of_realism}

VAEs utilise two probabilistic distributions during the generative process. A \textit{prior distribution} is assigned to generate random latent variables. Then, a neural network architecture (decoder) maps these to a conditional \textit{likelihood distribution}. A commonly used likelihood distribution is the multivariate Gaussian with a diagonal covariance matrix, which assumes independence among the output variables \cite{kingma2013auto}. While this assumption simplifies the modelling process, it often leads to poor sample quality. Specifically, by treating features as independent, VAEs tend to produce noisy data, as they fall outside the manifold of the original data \cite{Dorta_2018_CVPR}. This problem is particularly significant for time-series data generation, where capturing correlations between sequential data points is crucial for preserving informative patterns, such as peaks and cycles in energy consumption patterns. A common workaround to improve sample quality is to use the mean vector as the generated sample, which often results in overly smooth (``blurry") outputs. These two options for the generative process impose a significant limitation on VAEs' ability to generate realistic data, especially for applications that require fine-grained detail.

One natural question arises: Why not employ full covariance matrices to better capture feature dependencies? While full covariance matrices could, in theory, solve this issue, they introduce several practical challenges: 
\begin{itemize} 
    \item \textit{Parameterisation}: Full covariance matrices are more difficult to model as an output of a neural network, requiring elaborate mechanisms to ensure positive-definiteness.
    \item \textit{Singularity}: Generating full covariance matrices is prone to singularities, particularly during training, making them harder to constrain compared to their diagonal variants.
\end{itemize}
In \cite{Dorta_2018_CVPR} and \cite{langley2022structured}, these problems are addressed by using Cholesky and low-rank decompositions, respectively. However, these require either sparsity assumptions \cite{Dorta_2018_CVPR} or regularisation of the objective function \cite{langley2022structured}, which limits their representational power. Consequently, these challenges have limited the adoption of full covariance matrices in VAEs, leaving a gap in the literature and hindering progress toward generating high-quality, realistic synthetic data, particularly in fields that require accurate temporal or feature-based dependencies.

\subsection{Main Contributions}\label{sec:main_contributions}
In this work, we introduce \textbf{G}eneralized \textbf{U}ser-\textbf{I}nformed \textbf{D}ictionary \textbf{E}nhanced VAE (GUIDE-VAE), a novel framework that advances generative modelling for multi-user datasets through four main contributions: (i) a user-guided data generation process enabled by user embeddings, (ii) a novel covariance structure (PDCC) that enhances sample realism, and the applications of GUIDE-VAE for (iii) synthetic multi-user dataset generation and (iv) missing record imputation under data imbalance.

\begin{enumerate}[label=(\roman*)]
    \item \textbf{User embeddings for generative modelling:} We propose a novel methodology to condition generative models on user identities by employing user embeddings. LDA is used to create such embeddings by transforming users into vectors in a metric space that captures their similarities. These embeddings serve as conditions in a conditional generative model (in this work, a CVAE), enabling a \textit{user-guided} data generation process. To the best of our knowledge, this is the first instance of data-driven user information integration being applied to generative modelling.
        
    \item \textbf{Novel covariance structure and improved realism:} We introduce a new covariance matrix construction method, PDCC. It is applied in the GUIDE-VAE likelihood distribution to enhance sample realism. By learning \textit{dependency patterns} among features, PDCC effectively mitigates the noisy-sample issue commonly encountered in VAEs, while ensuring positive definiteness and avoiding singularity problems, without relying on sparsity or regularisation assumptions. This method maintains generalizability and provides a scalable solution for generating realistic data. To the best of our knowledge, this is the first time such a full covariance matrix parameterisation has been successfully implemented in this context.
    
    \item \textbf{Synthetic multi-user time series data generation:} We propose a novel approach to synthetic data generation that explicitly addresses the multi-user nature of datasets. GUIDE-VAE enables the generation of realistic time-series data across multiple users by conditioning on user embeddings, significantly improving modelling performance compared to conventional unguided CVAEs. This approach offers a new solution for generating diverse, realistic synthetic time-series data tailored to individual users, addressing a key limitation of current generative models by incorporating the multi-user aspect.

    \item \textbf{Inter-user information transfer and imputation:} We introduce the problem of missing record imputation in multi-user datasets under data imbalance, where the number of missing records varies among users. GUIDE-VAE leverages inter-user knowledge transfer through embeddings to mitigate this issue, utilising data from similar users to improve imputation accuracy. As a result, the model effectively handles users with fewer data points, naturally improving imputation performance without being explicitly designed to address data imbalance.

\end{enumerate}
    
We evaluated GUIDE-VAE using a multi-user energy consumption dataset to demonstrate its performance and the added benefits it offers. To simulate data imbalance, we created an enrollment model in which users enrol in the data collection system at random times, reflecting real-world conditions. In this use case, the dataset owner (e.g., a utility) aims to (1) create a synthetic version of the dataset and (2) complete the dataset by imputing the missing records for late-enrolling users. 
\section{Preliminaries}

\textbf{\textit{Notation.}} In the following, $\{\hat{x}_s\}\underset{S}{\overset{\sim}{\leftarrow}} p(x)$ means that $\{\hat{x}_s\}$ is a set of $S$ independent samples taken from $p(x)$. For a single sample, we similarly use $\hat{x}\overset{\sim}{\leftarrow} p(x)$. 

\subsection{Multi-user Time-series Profiles}\label{sec:profiling}

\begin{figure*}[t!]
    \centering
    \includegraphics[trim={0.5cm 0 0.5cm 0},clip,width=\textwidth]{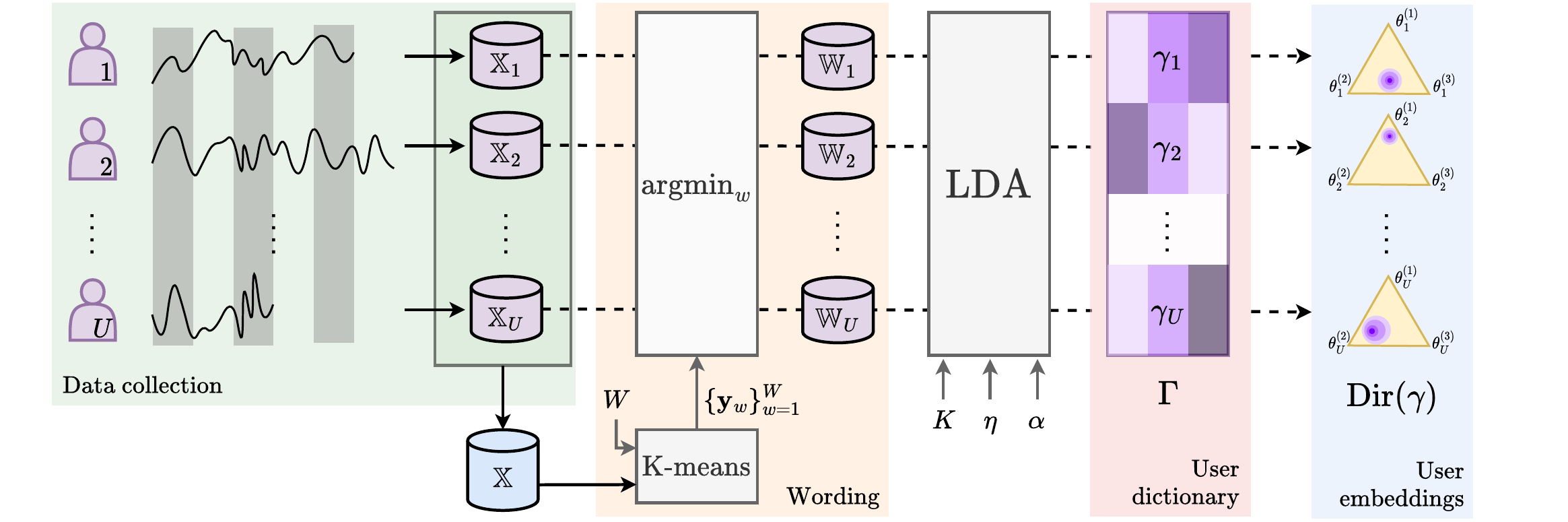}
    \caption{The user embedding framework for multi-user time-series datasets. Users' time-series data are segmented into profiles and clustered using the $k$-means clustering \cite{kaufman2009finding}. These clusters are treated as words, and user datasets are treated as documents for LDA. After training, LDA produces a user dictionary $\Gamma$, where each element corresponds to the parameters of a Dirichlet distribution, serving as user embeddings.}
    \label{fig:user_embedding_pipeline}
\end{figure*}

We consider datasets consisting of regularly and synchronously sampled time series from multiple sources (users in this study), as shown in Fig. \ref{fig:user_embedding_pipeline}. The data is segmented into profiles of length $T$, e.g., $T$=24 for daily profiles with hourly-sampled time series. We denote the $u^\text{th}$ user's dataset as $\mathbb{X}_u=\{\mathbf{x}_{un}\}_{n=1}^{N_u}$ where $\mathbf{x}_{un} = \left[ x_{un}^{(t)} \right]_{t=1}^T \in \mathcal{X}^{T} \subseteq \mathbb{R}^{T}$, where $N_u$ is the number of recorded profiles. These \textit{user datasets} can be pooled (anonymized) under $\mathbb{X} = \bigcup_{u=1}^U \mathbb{X}_u$ where $U$ is the total number of users. Note that each element $\mathbf{x}_i \in \mathbb{X}$ has a one-to-one correspondence to a user-indexed data point $\mathbf{x}_{un}$.\footnote{One mapping can be $i = n + \sum_{u'=0}^{u-1} N_{u'}, ~ N_0=0$.} We also assume that pieces of contextual information $\mathbf{c}_{un}$ are assigned to each data point  $\mathbf{x}_{un}$, which can be used as conditions for the generative modelling as described in the following sections.

\subsection{Latent Dirichlet Allocation} \label{sec:LDA}

LDA \cite{blei2003latent} is a variational method for topic discovery applied to collections of discrete data. It is essentially a multi-level Bayesian model that exploits the occurrence frequencies of data tokens (unique words) across data collections (documents) to cluster both data tokens and collections as mixture models. The mixture components are conventionally referred to as \textit{topics}, and a topic vector can be assigned to a given document using the posterior distribution of topics. In our context, LDA will allow us to discover latent structures in multi-user time-series data, enabling us to model user profiles as distributions over latent topics (user archetypes).

Let us consider a dataset consisting of $U$ documents, each consisting of $N_u$ words. The unique words $w_{un} \in \{1, \dots W \}$ are encoded as one of $W$ distinct values.\footnote{Conventionally, these values correspond to actual words. We keep them as integers for mathematical convenience. Note that the magnitude of these integer values does not convey any information.} This notation is used to represent documents $\mathbb{W}_u = \{ w_{un} \}_{n=1}^{N_u}$ and the entire dataset $\mathbb{W} = \{ \mathbb{W}_u \}_{u=1}^U$. Since this study utilises only the topics assigned to documents, not to tokens, the originally proposed Bayesian model in \cite{blei2003latent} can be reduced to a two-level hierarchy as the joint distribution
\begin{equation}
\begin{split}
    &p(\mathbb{W},\theta = \{\theta_u\}_{u=1}^U,\beta= \{\beta_k\}_{k=1}^K;\alpha,\eta)  \\ &=  {\color{OliveGreen}{p(\{\beta_k\};\eta)}} \prod_{u\in[U]}
    {\color{Bittersweet}{p(\mathbb{W}_u|\theta_u, \{\beta_k\})}} {\color{Violet}{p(\theta_u;\alpha)}}  \\ 
    &=  \prod_{k} {\color{OliveGreen}{\mathcal{D}\left(\beta_k;\eta\right)}} \times \prod_{u} {\color{Bittersweet}{\mathcal{M}(\mathbb{W}_u\Big| \sum_{k\in[K]} \theta_u^{(k)}\beta_k;N_u)}} {\color{Violet}{\mathcal{D}\left(\theta_u;\alpha\right)}}    
\end{split}
\end{equation}
where $K$ is the number of latent topics. Here, each $\theta_u \in \Delta^{K-1}$ represents the document-topic weights for the $u^\text{th}$ document and is \textcolor{Violet}{Dirichlet distributed} with concentration parameter $\alpha>0$. Similarly, the word weights for each topic $k$, $\beta_k \in \Delta^{W-1}$, is \textcolor{OliveGreen}{Dirichlet distributed} too, with the concentration parameter $\eta>0$. Lastly, each document $\mathbb{W}_u$ has a \textcolor{Bittersweet}{multinomial distribution} with $N_u$ number of trials and an event probabilities vector of $\sum_k^K \theta^{(k)}\beta_k \in \Delta^{W-1}$. 

Intuitively, this model offers a process for generating documents as bags of words from latent topics. In order to reverse this process and find the corresponding topic weights of a given document, one must acquire the posterior distribution $p(\theta_u|\mathbb{W}_u;\alpha, \eta) \propto \mathbb{E}_{{\color{OliveGreen}{p(\{\beta_k\};\eta)}}}\lbrack  {\color{Bittersweet}{p(\mathbb{W}_u|\theta_u, \{\beta_k\})}} \rbrack {\color{Violet}{p(\theta_u;\alpha)}}$ where the expectation is generally analytically intractable because the likelihood depends on $\sum_{k\in[K]} \theta_u^{(k)}\beta_k$, so the Dirichlet prior over $\beta_k$ is no longer conjugate to the multinomial. Instead, using variational inference, this posterior can be approximated as $q(\theta_u;\gamma_u) = \mathcal{D}(\theta_u;\gamma_u)$, where the parameter $\gamma_u$ can be determined using the expectation-maximisation algorithm in \cite{blei2003latent}. This variational inference from documents to the posterior parameters $\gamma_u$ is the core of the user embedding pipeline visualised in Fig. \ref{fig:user_embedding_pipeline}.

\subsection{Variational Autoencoders}\label{sec:vae}
VAEs \cite{kingma2013auto} are generative models that extend the variational Bayesian inference methodology to the realm of deep learning. They operate by using three key distributions: the likelihood $p_\psi(\mathbf{x}|\mathbf{z})$, the approximate posterior $q_\phi(\mathbf{z}|\mathbf{x})$, and the prior $p(\mathbf{z})$. Here, $\mathbf{z}$ is the latent variable (a learned representation) while $\mathbf{x}$ is the observed variable. Even though the family selection of these distributions is very flexible, it is common to choose multivariate normal distributions as follows:
\begin{align}
    p_\psi(\mathbf{x}|\mathbf{z}) &= \mathcal{N}(\mathbf{x}; \mu=f^\mu_\psi(\mathbf{z}), \Sigma = \text{diag}(f^\sigma_\psi(\mathbf{z})^2), \\
    q_\phi(\mathbf{z}|\mathbf{x}) &= \mathcal{N}(\mathbf{z}; \mu=f^\mu_\phi(\mathbf{x}), \Sigma = \text{diag}(f^\sigma_\phi(\mathbf{x})^2), \\
    p(\mathbf{z}) &= \mathcal{N}(\mathbf{z}; \mu = \mathbf{0}, \Sigma = \mathbf{I}).
\end{align}
Here, the neural networks $\{f^\mu_\psi,f^\sigma_\psi\}$ and $\{f^\mu_\phi,f^\sigma_\phi\}$ form the VAE's decoder and encoder neural networks, whose parameters are collected in $\psi$ and $\phi$, respectively. After setting up the distributions and neural networks, the VAE can be trained to maximise the evidence lower bound (ELBO) as
\begin{equation} \label{eq:elbo}
\begin{split}
    &\psi^*, \phi^*  \\ &= \argmax_{\psi, \phi} \sum_{\mathbf{x}_i \in \mathbb{X}}   \underbrace{\frac{\sum_l^{N^\text{MC}}\log p_\psi(\mathbf{x}_i|\hat{\mathbf{z}}_{il})}{N^\text{MC}} - \text{D}_\text{KL}(q_\phi(\mathbf{z}|\mathbf{x}_i)\lVert p(\mathbf{z}))}_{ \approx ~ \mathbb{E}_{q_\phi(\mathbf{z}|\mathbf{x}_i)} \left[ \log \frac{p_\psi(\mathbf{x}_i|\mathbf{z}) p(\mathbf{z})}{q_\phi(\mathbf{z}|\mathbf{x}_i)} \right] ~~(\text{ELBO})}
\end{split}
\end{equation}
where $\hat{\mathbf{z}}_{il} \overset{\sim}{\leftarrow} q_\phi(\mathbf{z}|\mathbf{x}_i)$ is the $l^\text{th}$ sample from the posterior distribution for a given data point $\mathbf{x}_i$, $N^\text{MC}$ is the number of Monte Carlo samples to approximate $\mathbb{E}_{q_\phi(\mathbf{z}|\mathbf{x}_i)}[\log p_\psi(\mathbf{x}_i|\mathbf{z})]$ and $\text{D}_\text{KL}(\cdot \lVert \cdot)$ is the Kullback-Leibler divergence.

Since the ELBO objective is a lower bound for the exact log-likelihood $\log p_\psi(\mathbb{X})$ \cite{kingma2013auto}, maximising it equips the VAE with two main functionalities. Firstly, one can take a sample from the marginal distribution model, $\hat{\mathbf{x}} \overset{\sim}{\leftarrow}  p_\psi(\mathbf{x})$, through ancestral sampling as $\hat{\mathbf{x}} \overset{\sim}{\leftarrow}  p_\psi(\mathbf{x}|\hat{\mathbf{z}})$, $\hat{\mathbf{z}} \overset{\sim}{\leftarrow}  p(\mathbf{z})$, since $p_\psi(\mathbf{x}) = \mathbb{E}_{p(\mathbf{z})} \left[ p_\psi(\mathbf{x}|\mathbf{z}) \right]$. This enables the generation of synthetic datasets. Secondly, the latent variables, conditioned on the observed data via the posterior distribution, can be used in various applications, such as dimensionality reduction and clustering. This study primarily focuses on the first functionality of the VAEs, and we invite researchers to investigate the effects of the proposed methods on the latent variables. 

\textbf{Conditional variational autoencoders}. The natural extensions of VAEs are their conditional counterparts, CVAEs \cite{sohn2015learning}. These generative models aim to model the conditional distribution $p_\psi(\mathbf{x}|\mathbf{c})$ rather than $p_\psi(\mathbf{x})$ as in conventional VAEs, as explained above. These conditions, $\mathbf{c}$, are generally known labels that partially specify the random variables $\mathbf{x}$. Thanks to this additional information, CVAEs can learn to generate data belonging to a given condition, $\hat{\mathbf{x}} \overset{\sim}{\leftarrow}  p_\psi(\mathbf{x}|\mathbf{c})$, which provides additional control on the synthetic data generation process.

\begin{figure*}[th!]
    \centering
    \includegraphics[trim={1.2cm 0 2.1cm 0},clip,width=\textwidth]{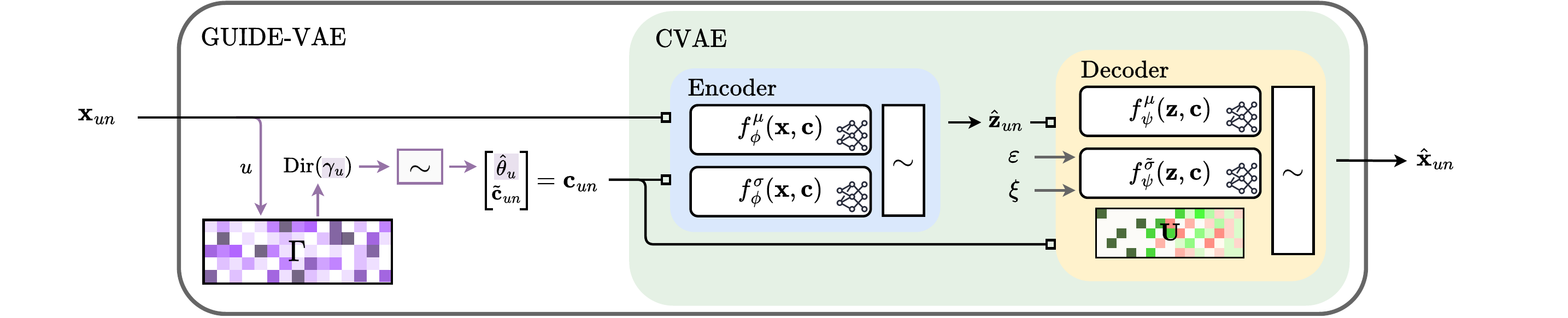}
    \caption{Overall computational diagram of GUIDE-VAE. GUIDE-VAE is a CVAE-based model enhanced with a learnable pattern dictionary for PDCC, which captures feature dependencies for improved realism. The user of the data point is selected from the user dictionary, and a sample from the corresponding probabilistic embedding is concatenated with auxiliary conditions (e.g., metadata or timestamps) and applied to both the encoder and decoder.}
    \label{fig:guide-vae_diagram}
\end{figure*}

Introducing the conditions into the VAEs requires only a minimal adjustment to the model and training. The derivation of ELBO remains the same, except for the structures of the distributions which take the forms $p_\psi(\mathbf{x}|\mathbf{z},\mathbf{c})$ and $q_\phi(\mathbf{z}|\mathbf{x},\mathbf{c})$.\footnote{Theoretically, the prior also takes the form of $p(\mathbf{z}|\mathbf{c})$, but it usually left untouched due to the extra inference load.} Thus, after concatenating the conditions into the observed and latent variables as inputs of the encoder and decoder networks, respectively, \eqref{eq:elbo} can be used for the CVAE training. The corresponding overall structure is depicted in Fig. \ref{fig:guide-vae_diagram}.\footnote{Note that this structure applies only to training. During the inference for synthetic data generation, one does not aim to reconstruct a data point but to generate it. This means applying ancestral sampling together with the conditions as $\hat{\mathbf{x}} \overset{\sim}{\leftarrow}  p_\psi(\mathbf{x}|\hat{\mathbf{z}},\mathbf{c}), ~ \hat{\mathbf{z}} \overset{\sim}{\leftarrow}  p(\mathbf{z})$.}

\section{Methodology}

The proposed GUIDE-VAE framework combines two core components: user embeddings extracted via LDA for user conditioning, and a CVAE enhanced with pattern dictionary-based covariance composition (PDCC) to capture feature dependencies and improve realism. Together, these components enable GUIDE-VAE to generate user-specific, high-quality time-series data.

\subsection{\textbf{Guidance}: LDA-based user-embedding}\label{sec:user_embedding}

As described in Section \ref{sec:LDA}, LDA assigns topic weights to each word in a vocabulary and uses these assignments to calculate the posterior probability distribution of topics for a given document. Therefore, after training, LDA can be used to map documents to Dirichlet distributions over topic mixtures. 
LDA's ability to infer latent topics from a collection of documents of variable length makes it an attractive candidate for inferring user behaviour from multi-user time series datasets. However, most time series datasets do not consist of words or discrete features in a finite domain, i.e. categorical data. Instead, they tend to have records in continuous domains. In \cite{chen2022constructing}, a data preprocessing pipeline is proposed to apply LDA to a non-textual time series dataset. This ``wording" process enables the continuous time series data to be mapped into a discrete space, making it compatible with LDA's document-word structure.

The wording consists of three steps, as depicted in Fig. \ref{fig:user_embedding_pipeline}. First, $k$-means clustering is applied to the pooled dataset $\mathbb{X}$ described in Section \ref{sec:profiling}. This results in $W$ cluster centres corresponding to a \textit{wording granularity} of $W$. Then, the user time series profiles $\mathbf{x}_{un}$ are replaced with their respective cluster labels, i.e. $w_{un} = \argmin_w \lVert\mathbf{x}_{un} - \mathbf{y}_{w}\lVert$ where $\mathbf{y}_{w} \in \mathcal{X}^{T}$ represents the $w^\text{th}$ cluster centre and $w \in \{1,\dots W\}$. Consequently, cluster labels derived from the profiles in user datasets are interpreted as words, and \textit{user documents} are formed as $\mathbb{W}_u = \{w_{un}\}_{n=1}^{N_u}$.

Substituting users for documents and user behavior for topics allows LDA to generate approximate posterior distributions for each user dataset, $\mathcal{D}(\theta_u;\gamma_u)$, where the concentration parameter $\gamma_u \in \mathbb{R}_+^K$ can be calculated iteratively using the document and the prior parameters as mentioned in Section \ref{sec:LDA}. The posterior distribution itself is considered \emph{probabilistic} user embedding, and the concentration parameter is its sufficient encoding. The concentration parameters of these embeddings are collected in a \textit{user dictionary} $\Gamma = \{ \gamma_u \}_{u=1}^U$ as shown in Fig. \ref{fig:user_embedding_pipeline}. 

Note that this differs from \cite{chen2022constructing}, which uses the mean of the posterior distribution, $\mathbb{E}\left[ \theta_u \right] = \frac{\gamma_u}{\sum_k \gamma_u^{(k)}}$ as the embedding. The reason to retain the distribution is that this conveys additional information. For instance, users with low data counts are assigned distributions with larger variances, reflecting the inherent uncertainty of topic allocation.

Embedding users as distributions introduces the challenge of applying these as conditions in mathematical models, such as CVAEs. To address this challenge, a sampling-based approach is employed in this study, where each distribution is represented by a collection of \textit{user vectors}, $\hat{\theta}_{us} \overset{\sim}{\leftarrow}  \text{Dir}(\gamma_u)$. This approach is not only computationally straightforward but also captures the variability and uncertainty inherent in user distributions, making it a desirable choice for large-scale datasets. In practice, if users are revisited indefinitely, as in neural network training, user vectors can be sampled per visit, and storing only the user dictionary $\Gamma$ is sufficient. Note that, for applications that require a single, concise representation, one can use the kernel mean embedding \cite{muandet2017kernel} of the Dirichlet distribution to map the user distribution to a deterministic vector. However, this is mathematically cumbersome and falls outside the scope of this study.

\subsection{\textbf{Realism}: Likelihood distributions using PDCC}

We propose a novel matrix composition method, termed PDCC, to parameterise the full covariance matrix of the GUIDE-VAE's likelihood distribution. This section provides the intuition behind PDCC, its theoretical foundation, and the useful properties it entails.

\subsubsection{Intuition}\label{subsec:intuition}
PDCC can be interpreted as a scheme in which random variables (features) are correlated by mapping high-dimensional, uncorrelated noise to a lower-dimensional space via fixed patterns. To elaborate, we aim to generate a rich noise distribution in $\mathbb{R}^T$. We do so by defining a set of unit vectors, $\{\mathbf{u}_v\in\mathbb{R}^T~\lvert~\lVert\mathbf{u}_v\lVert=1, v\in\lbrack V\rbrack\}$, that is larger in number than the dimension of the output space, i.e. $V>T$. Using this set, we can construct expressive multivariate noise patterns as linear combinations of these vectors, weighted by random coefficients, as
\begin{equation}
    \hat{\mathbf{\varepsilon}} =
    \sum_{v=1}^V \mathbf{u}_v \tilde{\sigma}^{(v)} \hat{\epsilon}^{(v)}  = \mathbf{U} \tilde{\Sigma}^{\frac{1}{2}} \hat{\epsilon}
\end{equation}
where $\tilde{\sigma}\in\mathbb{R}_+^V$, $\tilde{\Sigma}=\text{diag}(\tilde{\sigma})^2$, $\hat{\epsilon}\in\mathbb{R}^V$ and $\hat{\epsilon} \overset{\sim}{\leftarrow}\mathcal{N}(\mathbf{0},\mathbf{I})$. This construction naturally recovers standard noise sampling schemes as special cases. For example, if we choose $V = T$ and let the vectors $\mathbf{u}_v$ coincide with the canonical bases, then each feature is driven by its own independent noise source, yielding the familiar diagonal (independent) Gaussian noise. More generally, by allowing the vectors $\mathbf{u}_v$ to span arbitrary directions in $\mathbb{R}^T$ and combining them through the coefficients $\tilde{\sigma}^{(v)}$, the model induces structured correlations between features, effectively reproducing any desired covariance pattern. In this view, the vectors define directions of coupled variation, while the coefficients control their strength, jointly shaping the dependency structure. We choose $V \gg T$ to obtain an overcomplete set of patterns, which allows us to decouple scale from patterns and thereby increases the flexibility of the representation (as discussed later).

\subsubsection{Definition}
First, let $\tilde{\sigma} \in \mathbb{R}_+^V$ represent the \textit{auxiliary standard deviations} in a higher dimensional space ($V > T$). Next, let us define a transformation matrix \mbox{$\mathbf{U} = \left[ \mathbf{u}_1 \dots \mathbf{u}_V\right] \in \mathbb{R}^{T \times V}$}, where each column $\mathbf{u}_v \in \mathbb{R}^T$ has a unit norm, i.e. $\lVert \mathbf{u}_v \rVert = 1,~\forall v$. Lastly, let $\xi \in \mathbb{R}_+$ represent the \textit{base variance}. Having these three components, a covariance matrix $\Sigma \in \mathbb{R}_+^{T \times T}$ can be composed as 
\begin{equation}\label{eq:PDCC}
    \Sigma = \mathbf{U} \tilde{\Sigma} \mathbf{U}^\top + \xi\mathbf{I}
\end{equation}
where $\tilde{\Sigma} = \text{diag}(\tilde{\sigma})^2$. Note that the composition $\mathbf{U} \tilde{\Sigma} \mathbf{U}^\top$ is positive definite as long as $\text{rank}(U)=T$, and positive semi-definite otherwise, which can be proved easily using positive definiteness and matrix rank properties. In both cases, \eqref{eq:PDCC} ensures $\Sigma$ is always positive definite because $\xi>0$.

\subsubsection{Properties}
Motivated by the problem definition in \ref{sec:lack_of_realism}, we claim that PDCC provides the following properties.

\begin{itemize}
    \item \textbf{Constrainability} - PDCC is spectrally constrainable: Let $\{\tilde{\lambda}_t\geq0\}_{t=1}^T$ represent the set of eigenvalues of $\mathbf{U} \tilde{\Sigma} \mathbf{U}^\top$. Therefore, by the definition of eigenvalue decomposition, the eigenvalues of $\Sigma$ become $\{\lambda_t | \lambda_t = \tilde{\lambda}_t + \xi\}_{t=1}^T$, which makes them bounded from below by $\xi$. Meanwhile, likelihood calculations require inverting $\Sigma$, and this inversion fails if $\text{det}(\Sigma)\approx 0$ despite its positivity. The constant $\xi$ can prevent this singularity since \mbox{$\text{det}(\Sigma) = \prod_t \lambda_t = \prod_t (\tilde{\lambda}_t + \xi) \geq \xi^T$}. However, note that $\xi$ can be interpreted as an isotropic Gaussian noise (or density) added to the correlated patterns by default. To preserve the realism of the samples, this base variance must be kept at moderate levels. Unfortunately, $\xi^T$ diminishes exponentially with the increasing dimensionality, and one must sacrifice a certain level of realism to guarantee numerical stability in practice.

    \item \textbf{Flexibility} - PDCC is flexible enough to capture (linear) dependencies: Any positive definite covariance matrix $\Sigma \in \mathbb{R}^{T \times T}$ can be expressed as a sum of weighted rank-one components together with an isotropic term \cite{horn2012matrix}, which matches the structure induced by PDCC. Recalling Section \ref{subsec:intuition}, the columns $\{\mathbf{u}_v\}$ define directions of shared variation, while the coefficients $\tilde{\sigma}^{(v)}$ control their strength and $\xi$ captures a base variance, jointly allowing the construction of arbitrary covariance structures. Thus, when $V > T$, the parameterisation becomes overcomplete: while $\Sigma$ has $\frac{T(T+1)}{2}$ degrees of freedom, PDCC introduces more effective parameters (even under the unit-norm constraints on $\mathbf{u}_v$), implying that the same covariance matrix can be constructed in multiple ways.

    \item \textbf{Parameterizability} - PDCC is efficiently parameterizable for neural network training: We propose a partial parameterisation scheme for PDCC since its overparameterized nature makes it challenging to model it as an output of a neural network. In this scheme, only the auxiliary standard deviations $\tilde{\sigma}$ are related to latent variables using $f^{\tilde{\sigma}}_\psi(\mathbf{z})$ while the parameters of $\mathbf{U}$ are kept as global learnable parameters, i.e. as a single dictionary that is used to shape output noise for all generated samples. Lastly, the base variance $\xi$ is set as a hyperparameter to control the trade-off between stability and realism. Consequently, the resulting likelihood distribution is represented as
    \begin{equation}
        \begin{split}
        &p_{\psi,\mathbf{U}}(\mathbf{x}|\mathbf{z}) = \\
        &\mathcal{N}(\mathbf{x}; \mu=f^\mu_\psi(\mathbf{z}), \Sigma = \mathbf{U}\text{diag}(f^{\tilde{\sigma}}_\psi(\mathbf{z}))^2\mathbf{U}^\top + \xi\mathbf{I}).
        \end{split}
    \end{equation}
    
    One might question the benefit of having a dictionary of size larger than $T$, since the resulting covariance matrix still has rank $T$. However, because $\mathbf{U}$ is global, latent variables can influence the generated data only through the auxiliary standard deviations. This means that $\mathbf{U}$ must learn the most frequent patterns in the dataset, and $f^{\tilde{\sigma}}_\psi(\mathbf{z})$ activates the relevant patterns for the generated sample. Thus, the dictionary size $V$ allows controlling the permissible variety of fine-grained details in the generated data points. 
\end{itemize}

\subsection{GUIDE-VAE} \label{sec:overall_guide-vae}

The proposed GUIDE-VAE model is a CVAE enhanced with probabilistic user conditioning and PDCC, as depicted in Fig.~\ref{fig:guide-vae_diagram}. In contrast to standard CVAEs, which treat conditions as deterministic inputs, we model user information probabilistically. Specifically, for each user $u$, we define a latent user representation $\theta$ drawn from a user-specific prior distribution
\begin{equation}
    p(\theta \mid u) = \mathrm{Dir}(\gamma_u),
\end{equation}
where $\gamma_u$ is obtained from the LDA-based user embedding described in Section~\ref{sec:user_embedding}. Thus, the conditional likelihood of a data point $\mathbf{x}$ given user $u$ involves marginalisation over $\theta$:
\begin{equation}\label{eq:guide-vae_dist}
\begin{split}
    p(\mathbf{x} \mid u, \tilde{\mathbf{c}}) &=
    \int p_{\psi,\mathbf{U}}(\mathbf{x} \mid \mathbf{z}, \theta, \tilde{\mathbf{c}})
    p(\mathbf{z}) p(\theta \mid u)
    \, d\mathbf{z} \, d\theta \\
    &\approx \frac{1}{S^\text{MC}}\sum_{s=1}^{S^\text{MC}} \int p_{\psi,\mathbf{U}}(\mathbf{x} \mid \mathbf{z}, \mathbf{c} = \left[\hat{\theta}_{us}, \tilde{\mathbf{c}}\right])
    p(\mathbf{z})
    \, d\mathbf{z} .
\end{split}
\end{equation}
Here, the first integrand is the likelihood distribution parameterised via PDCC,
\begin{equation}\label{eq:guide-vae_likelihood}
    p_{\psi,\mathbf{U}}(\mathbf{x} \mid \mathbf{z}, \mathbf{c}) =
    \mathcal{N}\left(
    \mathbf{x};
    f^\mu_\psi(\mathbf{z}, \mathbf{c}),
    \mathbf{U}\mathrm{diag}(f^{\tilde{\sigma}}_\psi(\mathbf{z}, \mathbf{c}))^2\mathbf{U}^\top + \xi\mathbf{I}
    \right),
\end{equation}
$\tilde{\mathbf{c}}$ contains auxiliary contextual variables, e.g. calendar variables, and
$\hat{\theta}_{us} \overset{\sim}{\leftarrow} p(\theta | u)$. In this study, we set the number of MC samples for user embeddings to $S^\text{MC}$=1 for ease of computation, in a similar spirit to \cite{kingma2013auto}. The remaining integral is identical to a conventional CVAE objective and can be trained using the same ELBO given in Section \ref{sec:vae}. Note that we preserve the distributional representation of the embeddings by sampling a new $\hat{\theta}_{us}$ every time the datapoint $\mathbf{x}_{un}$ is re-visited during the neural network training.

Furthermore, we apply a lower bound constraint on the auxiliary standard deviations in \eqref{eq:guide-vae_likelihood}, i.e. 
$\forall v : f^{\tilde{\sigma}^{(v)}}_\psi(\mathbf{z}, \mathbf{c}) > \varepsilon $, to reinforce numerical stability and prevent degenerate solutions.
Without this constraint, the decoder may disregard the pattern dictionary, causing the likelihood to collapse to an isotropic Gaussian
$\mathcal{N}(\mathbf{x}; f^\mu_\psi(\mathbf{z}, \mathbf{c}), \xi\mathbf{I})$.
Enforcing this constraint ensures that PDCC actively contributes to modelling feature dependencies and learns informative correlation structures.
\section{Experiments}

\subsection{Data preperation}

\subsubsection{Dataset}
The dataset used in this study consists of smart meter data collected from various electricity consumers across 47 provinces in Spain, including homes, offices, and businesses \cite{quesada2024electricity}. The dataset comprises 25,559 customers (users) and their hourly electricity consumption measurements (in kWh) collected between November 2014 and June 2022.\footnote{The dataset has two versions: raw with missing values and the imputed one. This study uses the imputed version.} This dataset naturally contains an imbalance between users in terms of data quality, because consumers enrolled in the data collection system at different times.

This study utilises a spatiotemporal subset of the entire dataset, specifically the data collected from Gipuzkoa (the province with the highest data density) between June 2021 and June 2022, for the experiments. For a controlled experiment, only users with at least 1 full year of enrollment from June 2021 to June 2022 were considered, and the experiments were conducted for this period. Lastly, users who consistently consume 0~kWh or exhibit negative consumption at least once are also eliminated. These resulted in a dataset with $U$=6830 users and $N_u$=365 daily ($T$=24) profiles per user, totalling $\sim$2.5M records.

One of the motivating challenges for the GUIDE-VAE approach is imputation and improved forecasting performance under data imbalance. This problem is already apparent in the data collection in \cite{quesada2024electricity}, and the sub-dataset created above can be used to simulate the late enrolment problem. Note that this crafted dataset has ground truths for the missing records, unlike the original one.

\begin{figure}
    \centering
    \includegraphics[width=0.95\linewidth]{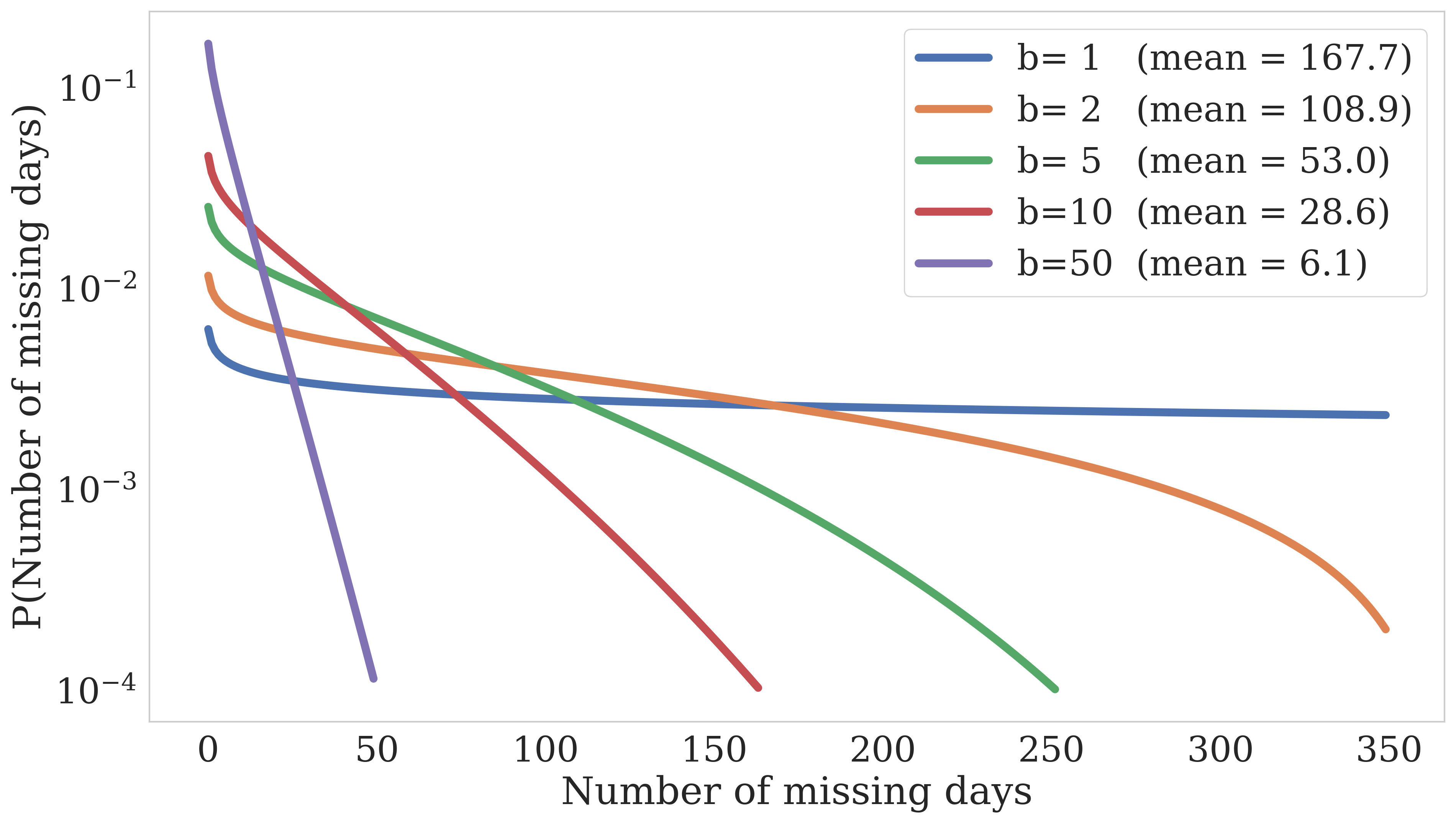}
    \caption{The pmf of beta-binomial distribution ($n$=365, $a$=0.85) in logarithmic scale for different $b$ values.}
    \label{fig:betabinom}
\end{figure}

\begin{figure}
    \centering
    \includegraphics[width=\linewidth]{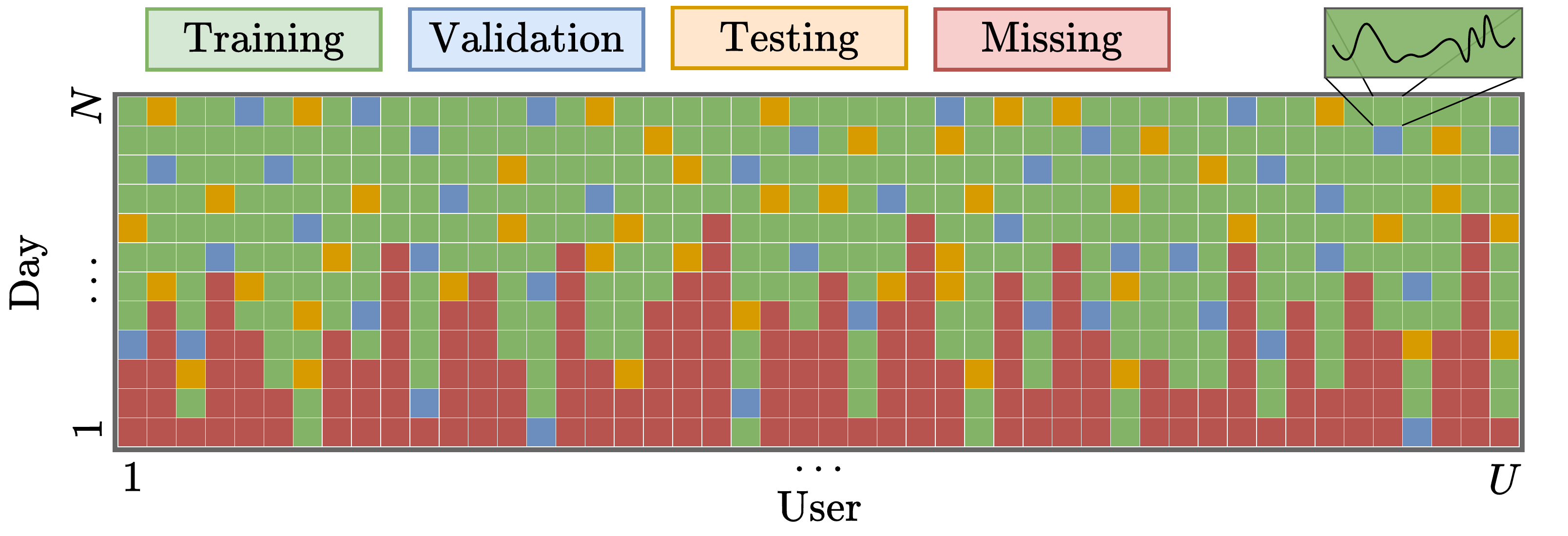}
    \caption{The data splitting scheme used in experimentation. A full dataset in which each user has an equal number of profiles is first amputated according to a beta-binomial distribution, and the resulting profiles are reserved in the missing set (in red). The remaining were randomly split into training (green), validation (blue), and testing (yellow) sets.}
    \label{fig:enrolment_matrix}
\end{figure}

In order to create artificial missingness, the data of each user's first days are ``amputated" randomly, corresponding to late enrolments. For this purpose, the beta-binomial distribution with the probability mass function (pmf)
\begin{equation}
    P(M=m;a,b,n) = \begin{pmatrix} n \\ m \end{pmatrix} \frac{\mathcal{B}(m-a, m-k+b)}{\mathcal{B}(a, b)}
\end{equation}
is employed to independently sample the employment date for each user, i.e. $N_u = 365-M_u$ where $M_u \sim \text{BetaBinom}(a,b,n),~ \forall u$. Here, $\mathcal{B}$ is the Beta function, $a>0$ and $b>0$ are shape parameters, and $n$ represents the length of the integer support $m\in\{0,1,\dots n\}$. In this study, the parameters $n$ and $a$ are set to 365 and 0.85, respectively; thus, the \textit{severity of missingness} is controlled by the parameter $b$. In Fig. \ref{fig:betabinom}, the pmfs generated by different $b$ values are plotted on a logarithmic scale. The mean enrolment day is given by $\mathbb{E}\left[M\right]=\frac{na}{a+b}$, which decreases with increasing $b$; hence $b$ is referred to as the \textit{data availability parameter}. An illustration of simulated missingness due to late enrolment is given in Fig. \ref{fig:enrolment_matrix}.

\subsubsection{Preprocessing}
\begin{figure}[!t]
\centering
\subfloat[]{\includegraphics[height=0.427\linewidth]{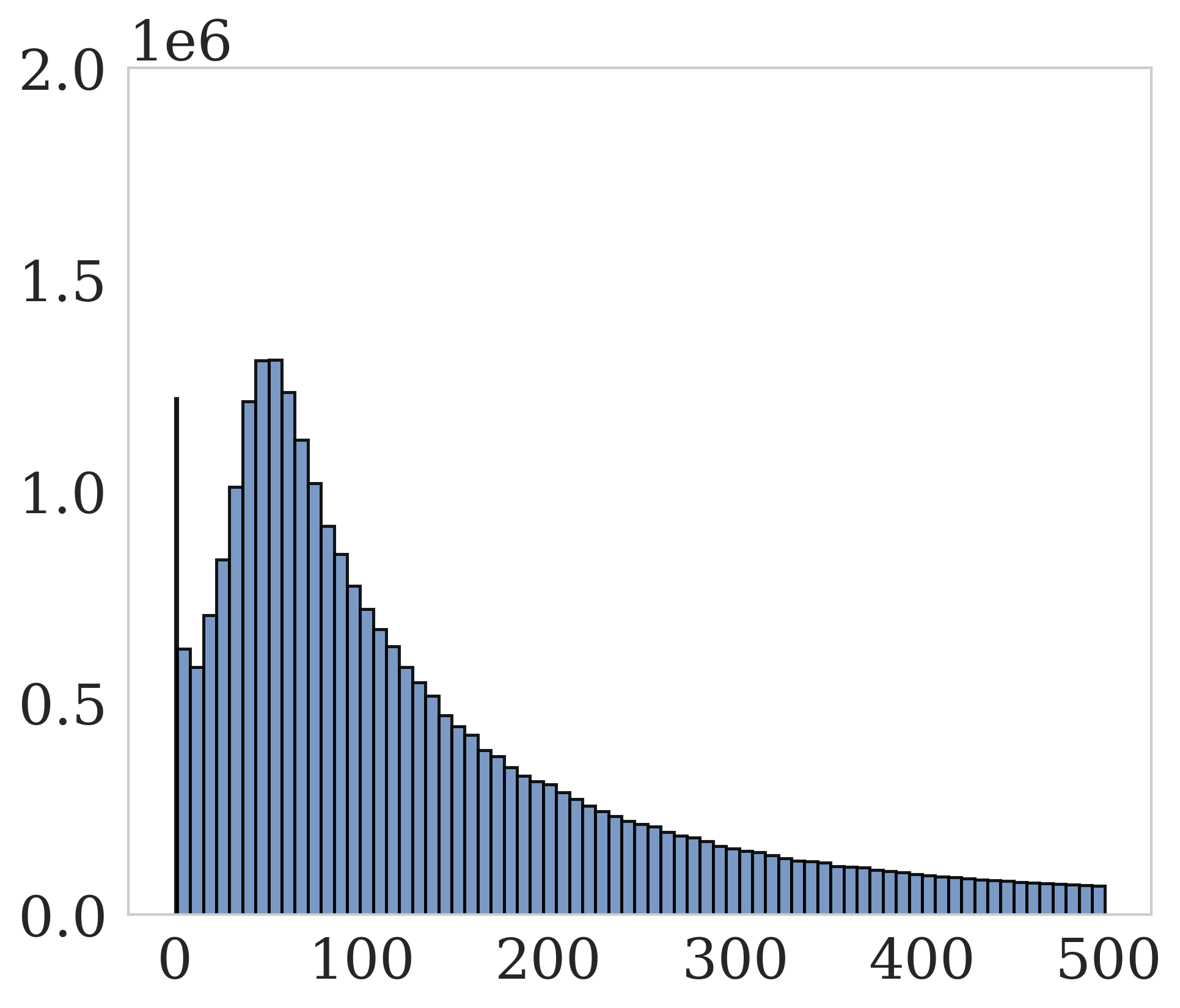}\label{fig:unnormalized_histogram}}
\hfill
\subfloat[]{\includegraphics[height=0.405\linewidth]{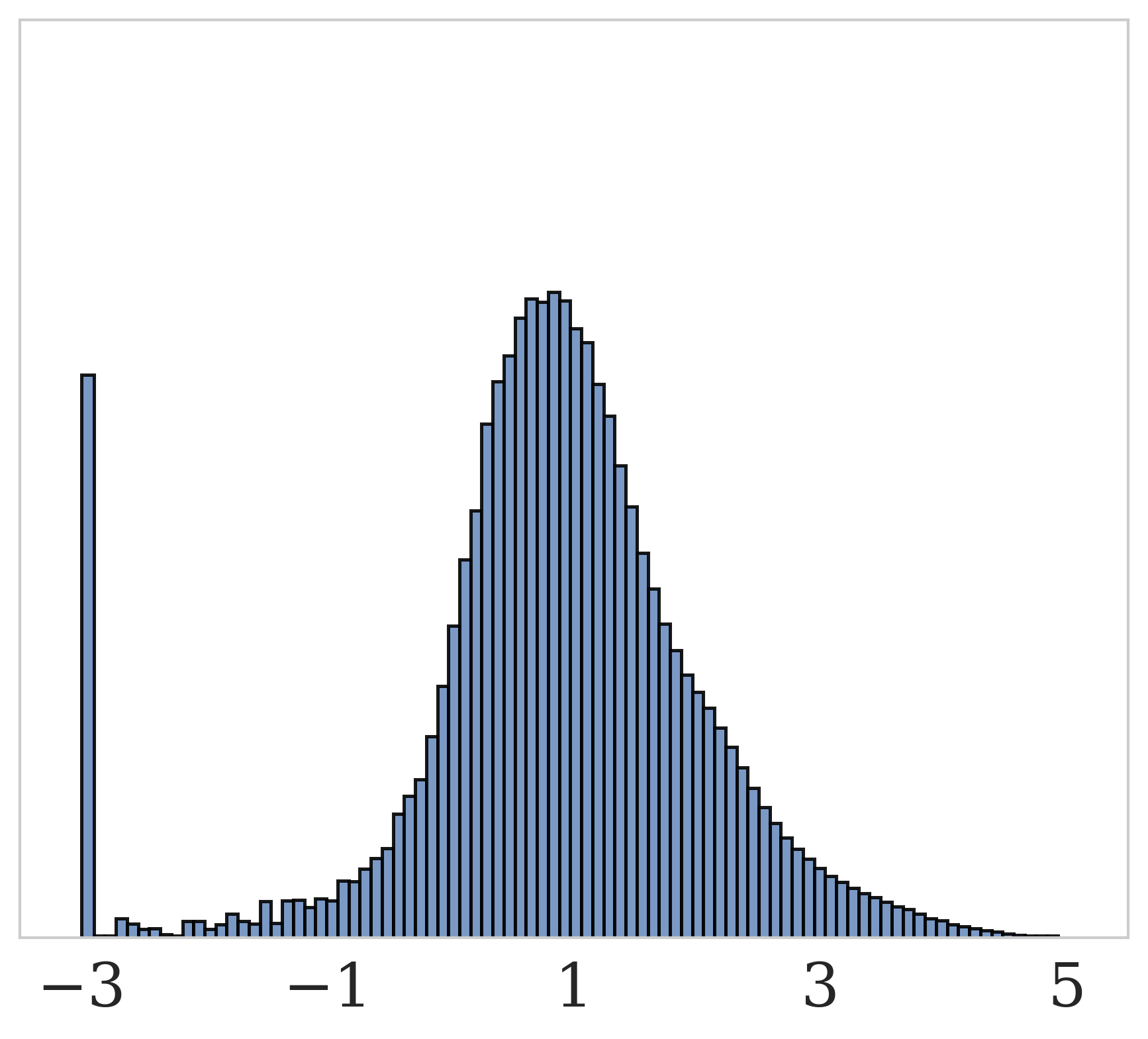}\label{fig:normalized_histogram}}
\label{fig:histograms}
\caption{The distributions of (a) measurements (in Wh) in the dataset and (b) the zero-preserved log-normalisation ($h_0$=-3, $h_+$=1) applied to them.}
\end{figure}

Energy consumption data often exhibits occasional peaks, leading to a heavy-tailed distribution, as illustrated in Fig. \ref{fig:unnormalized_histogram}. The conventional approach to handling such heavy-tailed data is by applying a logarithmic transformation. However, energy consumption data is inherently zero-inflated, meaning that there are data instances (in time) that have no energy consumption. These exact zero values prevent the direct use of the logarithmic transformation. To mitigate this, we propose a novel scaling technique called zero-preserved log-normalisation.

\begin{definition}[Zero-preserved log-normalization]
    Given a non-negative scalar dataset $\mathbb{X} = \{x_i\in \mathbb{R}_{\geq 0}\}_{i=1}^N$, two index sets can be created as $\mathbb{I}_+=\{i| x_i>0, x_i\in\mathbb{X} \}$ and  $\mathbb{I}_0=\{i| x_i=0, x_i\in\mathbb{X} \}$, representing the indices of positive and zero valued elements in $\mathbb{X}$, respectively. Zero-preserved log-normalization transforms the dataset $\mathbb{X}$ into $\bar{\mathbb{X}} = \{\bar{x}_i\}$ where 
    \begin{equation} \bar{x}_i = f^{\text{ZPLN}} (x_i; m, s, h_+, h_0 )= \begin{cases} 
      \frac{\log x_i - m}{s}  + h_+, & i\in\mathbb{I}_+ \\
      h_0, & i\in\mathbb{I}_0  \end{cases}.
    \end{equation}
    Here, $m=\frac{1}{\lvert\mathbb{I}_+\rvert} \sum_{i\in\mathbb{I}_+} \log x_i$, $s^2 = \frac{1}{\lvert\mathbb{I}_+\rvert - 1} \sum_{i\in\mathbb{I}_+} (\log x_i - m)^2 $. Constants $h_0$ and $h_+$ ($h_+>h_0$) are chosen offsets. 
\end{definition}

The constants $h_0$ and $h_+$ are used to provide an effective separation between zero and positive consumption values.\footnote{Please note that the zero-preserved log-normalization differs from the conventional offsetting ($\bar{x}_i = x_i + \delta, ~ \forall i\in [N]$) and zero-replacement ($\bar{x}_i = \delta, ~ \forall i\in \mathbb{I}_0$) by excluding zeros from the estimation of $m$ and $s$, so that they are not affected by unwanted skewness caused by the zeros.} The resulting data distribution after applying zero-preserved log-normalization, with $h_0$=-3 and $h_+$=1, is shown in Fig. \ref{fig:normalized_histogram}, where the desired log-normality can easily be seen. Also, choosing $h_0$ sufficiently smaller than $h_+$ makes the inverse transformation robust to small deviations around $h_0$, i.e. if $\bar{x}=h_0\pm\delta$ and $h_+ \gg h_0 + \frac{m}{s} +\delta$, then $x=\exp (s(\bar{x}-h_+) + m) \approx 0 $. 

The resulting preprocessing flow is as follows. First, the dataset is created for a given availability parameter $b$, and missing records are stored in $\mathbb{X}^{\text{missing}}$. Then, the remaining data is split into training, validation, and test sets with a ratio of 8:2:2, whose sizes also depend on the selected $b$. This splitting process is illustrated in Fig. \ref{fig:enrolment_matrix}. Then, the zero-preserved log-normalization ($h_0$=-3, $h_+$=1) is applied to each feature, $x^{(t)}$, individually. Note that the zero-excluded mean $m^{(t)}$ and standard deviations $s^{(t)}$ are estimated using only the training set but applied to all sets.  

\subsubsection{Conditioning}

As mentioned in Section \ref{sec:overall_guide-vae}, there are two main components of the conditions ($\mathbf{c}_{un}$) used in the GUIDE-VAE training: the sampled user vector $\hat{\theta}_u$, and the auxiliary condition vector $\tilde{\mathbf{c}}_{un}$. The generation of the user dictionary $\Gamma$ is explained thoroughly in Sections \ref{sec:LDA} and \ref{sec:user_embedding}. Similar to the normalisation step, only the training data is used in LDA-based user embedding, and the resulting dictionary $\Gamma$ is stored for a given user embedding hyper-parameter combination ($W$, $K$). The prior parameters are set to $\eta=\frac{1}{W}$ and $\alpha = \frac{1}{K}$.

Two auxiliary conditions were used: months and weekdays. These conditions are encoded using the cyclic ($\sin$-$\cos$) transformation as in \cite{wang2022contextual}, and assigned to their respective data points $\mathbf{x}_{un}$. Due to the intrinsic seasonality of the data, we believe these features (or those that encode similar information) are essential for energy consumption time series modelling.\footnote{Since the aim is not finding the best generative model but showcasing the benefits of GUIDE-VAE, the number of auxiliary conditions is limited to two.}

\subsection{Training}
\subsubsection{Neural networks}
\begin{figure}
    \centering
    \includegraphics[width=0.8\linewidth]{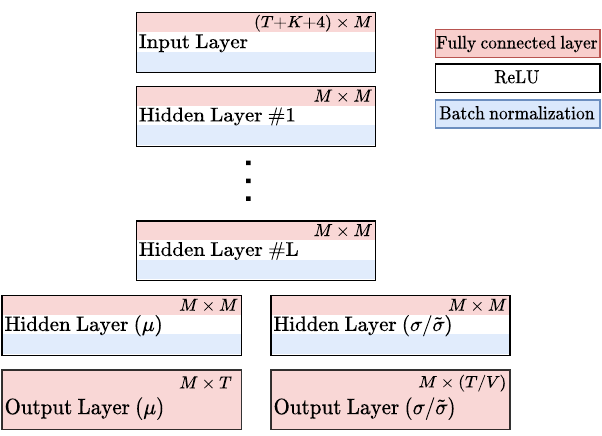}
    \caption{The neural network architecture used both for the encoder and the decoder.}
    \label{fig:nn_architecture}
\end{figure}

The neural network structure used in the experiments, both for the encoder and decoder, is depicted in Fig. \ref{fig:nn_architecture}. The input size $(T+K+4)$ is the same for both networks, because the size of the latent space is fixed at $T$=24 in all experiments, as the main objective is data modelling, not dimensionality reduction. The remaining part of the input $(K+4)$ comes from the conditions. The only difference between the two networks is the output size of the Output Layer ($\sigma/\tilde{\sigma}$), which equals $T$ for the encoder and $V$ for the decoder. Lastly, $L$ and $M$ are set to 3 and 1000, respectively, for all experiments.

\subsubsection{Constraints}
There are various constraints on the outputs of both the encoder and the decoder to ensure numerical stability. Firstly, the base variance $\xi$ is set to $10^{-2}$, which means the marginal standard deviations of the likelihood distribution are bounded below by $10^{-1}$. Similarly, this bound for the posterior distribution is set to $5\times10^{-1}$. Likewise, the mean parameters of both distributions are constrained between -3 and 5, resulting in better convergence without compromising the modelling performance.

\subsubsection{Optimisation}
Adam \cite{kingma2014adam} is used for the optimisation of the GUIDE-VAE with its default parameters ($\beta_1$=0.9, $\beta_2$=0.999), and L2-regularisation is applied over all parameters (including $\mathbf{U}$) with a coefficient of $10^{-5}$. Moreover, an adaptive learning rate scheduler and early stopping are applied using the validation set to further prevent overfitting. Lastly, $N^\text{MC}$=16 is used for Monte Carlo sampling in the ELBO \eqref{eq:elbo}.

\subsection{Performance metric}
In order to assess the performance of the trained generative models, two datasets are utilised: $\mathbb{X}^{\text{test}}$ for synthetic data generation and $\mathbb{X}^{\text{missing}}$ for missing-record imputation. The log-likelihood of a given set on the generative model
\begin{equation}
    \mathbb{E}_{p^{\cdot}(\mathbf{x},u,\tilde{\mathbf{c}})} \left[ \log p (\mathbf{x}|u,\tilde{\mathbf{c}}) \right] = \frac{1}{\lvert \mathbb{X}^{\cdot} \rvert} \sum_{\mathbf{x}_{un}\in \mathbb{X}^{\cdot}} \log p (\mathbf{x}_{un}|u,\tilde{\mathbf{c}}_{un})
\end{equation}
is chosen as the performance metric. Here $p^{\cdot}(\mathbf{x},u,\tilde{\mathbf{c}})$ is the empirical distribution representing either the testing or missing datasets and their respective conditions. Since $\log p (\mathbf{x}_{un}|u,\tilde{\mathbf{c}}_{un})$ is intractable, it is approximated by importance sampling as
\begin{equation}
\begin{split}
     &\log p (\mathbf{x}_{un}|u,\tilde{\mathbf{c}}_{un}) =  \log \mathbb{E}_{q}\left[\frac{p(\mathbf{x}_{un},\mathbf{z}|u,\tilde{\mathbf{c}}_{un})}{q(\mathbf{z}|\mathbf{x}_{un},u,\tilde{\mathbf{c}}_{un})} \right] \\
    &\approx  -\log S^{\text{MC}} + \log \sum_{s=1}^{S^{\text{MC}}}  \frac{p_\psi(\mathbf{x}_{un}|\hat{\mathbf{z}}_{uns},\hat{\theta}_{us},\tilde{\mathbf{c}}_{un}) p(\hat{\mathbf{z}}_{uns})}{q_\phi(\hat{\mathbf{z}}_{uns}|\mathbf{x}_{un}, \hat{\theta}_{us},\tilde{\mathbf{c}}_{un})}
\end{split}
\end{equation}
where $\{\hat{\mathbf{z}}_{uns}\} \underset{S^{\text{MC}}}{\overset{\sim}{\leftarrow}} q_\phi(\mathbf{z}|\mathbf{x}_{un}, \hat{\theta}_{us},\tilde{\mathbf{c}}_{un})$ and $\{\hat{\theta}_{us}\} \underset{S^{\text{MC}}}{\overset{\sim}{\leftarrow}} \text{Dir}(\gamma_u)$. After sweeping different values, we set $S^{\text{MC}}$=100 for all experiments to balance computational time and estimation variance.

Note that log-likelihood-based performance assessment is considerably more appropriate for the assessment of high-dimensional distributions than conventional sample comparison-based performance metrics, which require a large sample population to accurately represent the distributions being compared. In this case, the size of the original dataset quickly becomes limiting. For instance, the dataset used in this study has three conditions: users, months, and weekdays. This means there are at most five instances of a given condition triplet in a year-long dataset, and they are far from effectively representing a conditional distribution. This problem gets even more severe when continuous conditions are involved.

\section{Results \& Discussion}

The results of the experiments conducted are presented here. Unless otherwise stated, the following hyperparameter values were used as defaults:
\begin{itemize}
    \item Data availability: $b=10$
    \item User vector size: $K=100$
    \item Pattern dictionary size: $V=100$
    \item Wording granularity: $W=1000$
    \item Auxiliary standard deviation lower-bound: $\varepsilon=10^{-4}$
    \item Base variance: $\xi=10^{-2}$
\end{itemize}
These values were selected based on preliminary tuning to strike a balance between computational efficiency and performance.

\subsection{Visual inspection}

\begin{figure*}[htbp!]
    \centering
    \includegraphics[width=\linewidth]{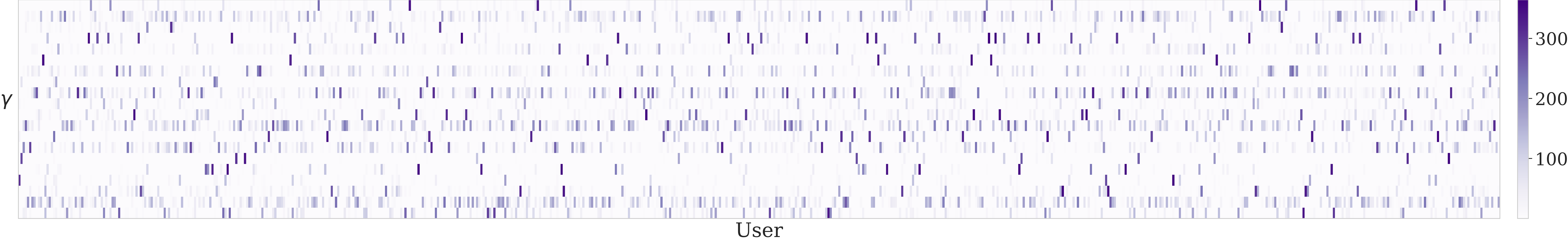}
    \caption{A learned user dictionary $\Gamma$ with $K$=20. Each column corresponds to the concentration parameter ($\gamma_u$) of a Dirichlet distribution representing a user.}
    \label{fig:user_gamma}
\end{figure*}

The user and pattern dictionaries, as well as the generated time series samples after training, are visualised in this section. 
Fig. \ref{fig:user_gamma} shows a 20-dimensional ($K$=20) user dictionary $\Gamma$ for 10\% of the users. The diversity of patterns suggests that the method distinguishes users based on a variety of observed behaviours.

\begin{figure*}[htbp!]
\centering
\subfloat[]{\includegraphics[height = 90pt ]{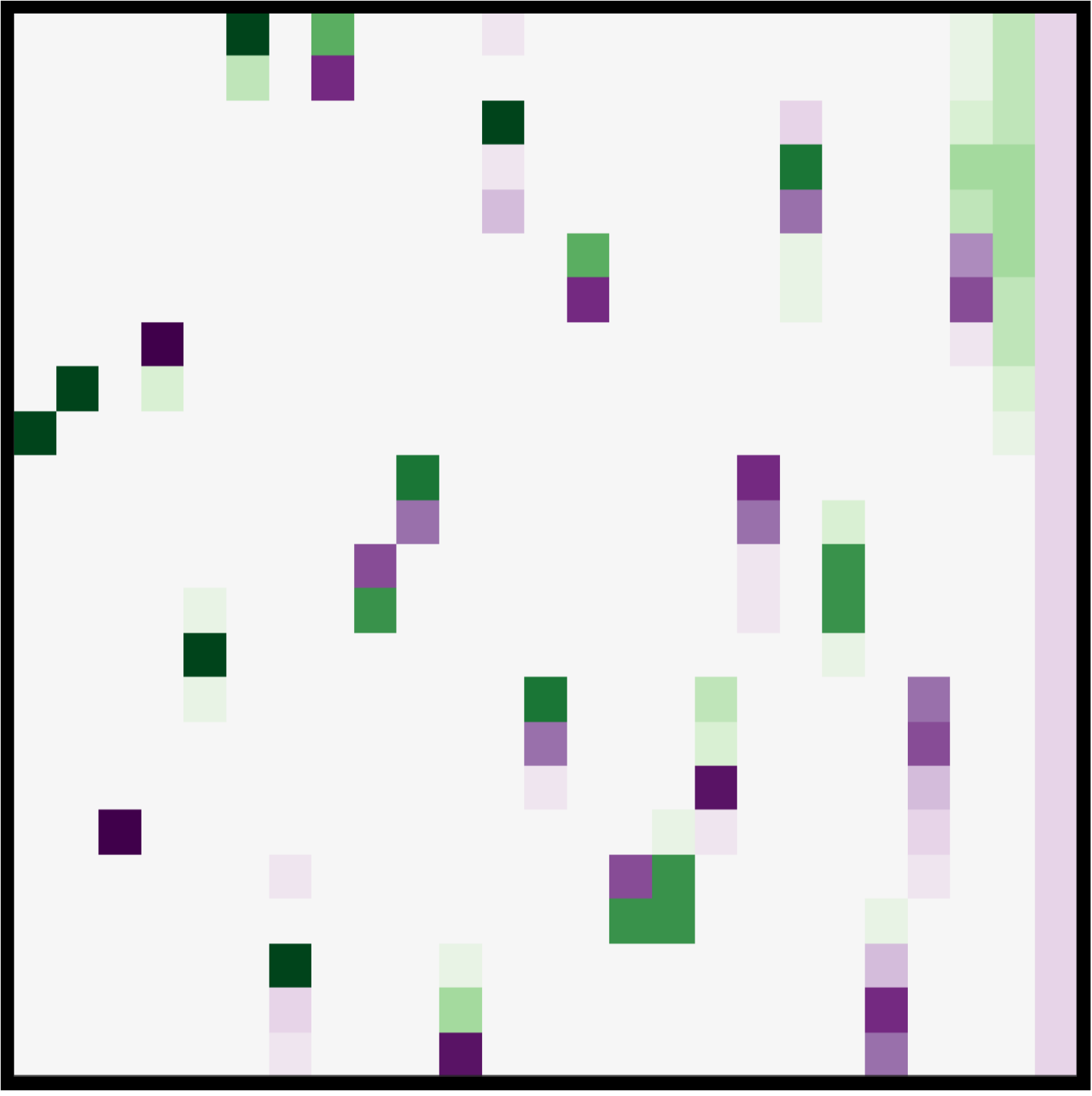}
\label{fig:pattern_dictionary_sorted_25}}
\hfil
\subfloat[]{\includegraphics[height = 90pt ]{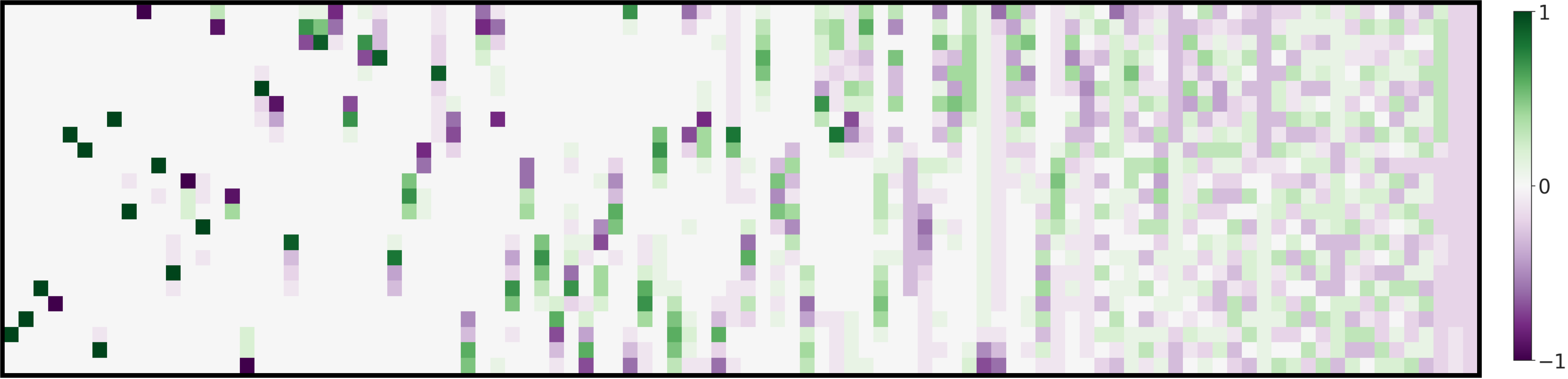}
\label{fig:pattern_dictionary_sorted_100}}
\caption{Pattern dictionaries $\mathbf{U}$ from two trained GUIDE-VAE models with (a) $V$=25 and (b) $V$=100, sorted according to $L_1$-norm of the pattern vectors.}
\label{fig:pattern_dictionary_sorted}
\end{figure*}

Next, we investigate the pattern dictionaries of two GUIDE-VAE models with $V$=25 and $V$=100, shown in Fig. \ref{fig:pattern_dictionary_sorted}. To enhance the presentation, the pattern vectors ($\mathbf{u}_v$) are sorted with respect to their $L_1$-norm $\rVert\mathbf{u}_v\lVert_1$ (decreasing), placing sparse patterns on the left and dense patterns on the right. The richer pattern selection available in the larger dictionary is immediately apparent and expected, as it allows GUIDE-VAE to exploit finer details and temporal correlations. For instance, a large variety of switching behaviours are captured by the model on the right side of the dictionary in Fig. \ref{fig:pattern_dictionary_sorted_100}, which are absent in Fig. \ref{fig:pattern_dictionary_sorted_25}. Another noticeable aspect is that many patterns have adjacent green (positive) and purple (negative) patches, corresponding to energy consumption shifting in time.

\begin{figure*}[t!]
    \centering
    \includegraphics[width=\linewidth]{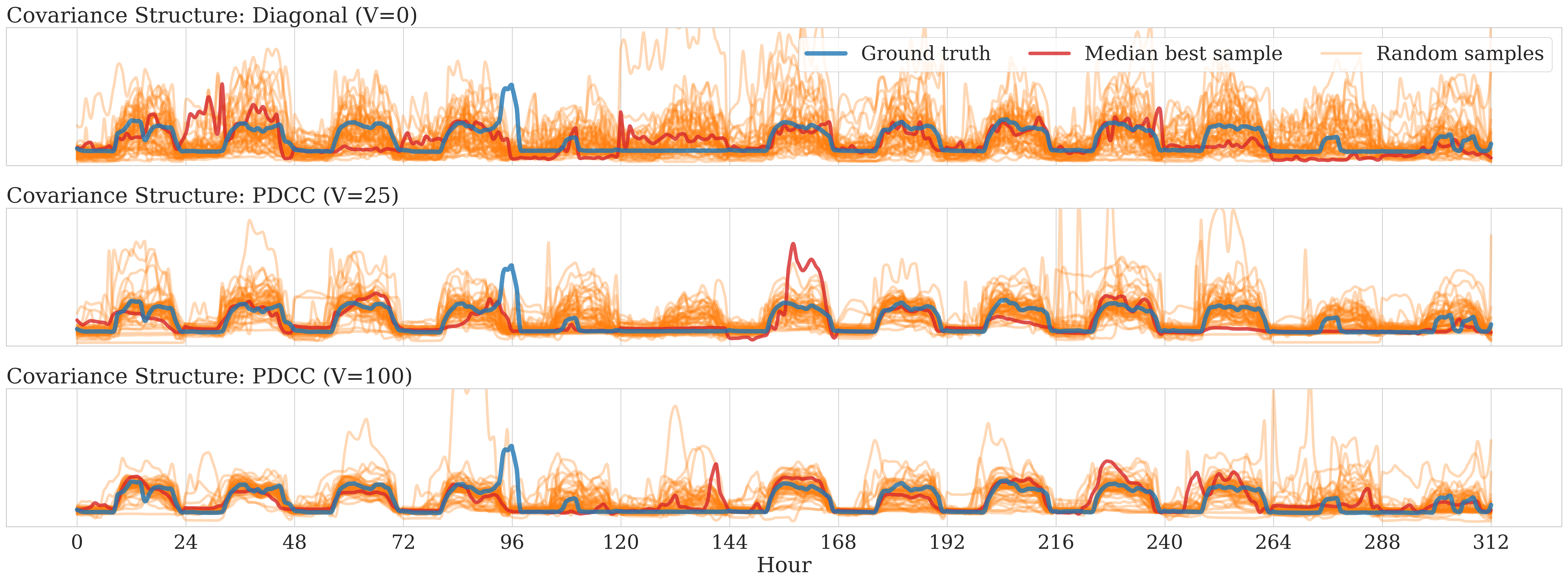}
    \caption{The effect of pattern dictionary size $V$ on the sample quality. A selected user's missing time series measurements (in blue) and the median-best imputation (in red) out of 10000 generated time series (50 of which are in orange). $V$=0 corresponds to vanilla diagonal covariance matrix modelling.}
    \label{fig:sample_comparison}
\end{figure*}

Lastly, we visualise the data generated by GUIDE-VAE with varying dictionary sizes. For this, we conducted a sub-experiment in which we selected a random user $u'$ from the dataset, and we imputed its missing time series, $\mathbb{X}^\text{missing}_{u'} = \mathbb{X}^\text{missing} \cap \mathbb{X}_{u'}$, $10^4$ times by sampling $\{\hat{\mathbf{z}}_{ns}\}\underset{10^4}{\overset{\sim}{\leftarrow}} p(\mathbf{z})$ for all days $n$ and $\{\hat{\mathbf{\theta}}_{s}\} \underset{10^4}{\overset{\sim}{\leftarrow}} \text{Dir}(\gamma_{u'})$. Note that each sample $s$ from the prior distribution transforms to a likelihood distribution in the data space, and we aim to visualize the samples from these likelihood distributions, $\hat{\mathbf{x}}_{ns} \overset{\sim}{\leftarrow} p_\psi(\mathbf{x}_n|\hat{\mathbf{z}}_{ns}, \hat{\theta}_{s}, \tilde{\mathbf{c}}_n)$, since these are the final outcomes during inference. For visualization we concatenated the profiles of a given sample $s$ as $\hat{\mathbf{x}}_s = \left[ \hat{\mathbf{x}}_{ns} \right]_{n=1}^{M_{u'}} \in \mathcal{X}^{M_{u'} T }$, where $\big \lvert \mathbb{X}^\text{missing}_{u'} \big \rvert = M_{u'}$. The resulting samples for each model are given in Fig. \ref{fig:sample_comparison}. In order to showcase the typical performance of each model, we calculated the log-likelihood of the ground truth time series on each sampled distribution, i.e. $\sum_{\mathbf{x}_{n}\in\mathbb{X}^\text{missing}_{u'}} \log p_\psi(\mathbf{x}_{n}|\hat{\mathbf{z}}_{ns}, \hat{\theta}_{s}, \tilde{\mathbf{c}}_n), ~ \forall s$, and stored the sample that corresponds to median of these scores. This ``median best sample" is also given in Fig. \ref{fig:sample_comparison}.

As shown in Fig. \ref{fig:sample_comparison}, all models can generate samples that are very close to the ground truth. Please note that these were generated using only the user embedding, month, and weekday information pieces. Also, recall that the ground-truth profile was never seen during training, indicating GUIDE-VAE's generalisation ability. Another qualitative outcome of this experiment is the plausibility of the generated samples. As one can see, not only the median samples but all samples become less noisy as the pattern dictionary size increases. This suggests that the larger the pattern dictionary, the more likely it is to observe realistic samples.

\subsection{Effect of user embeddings} 
\begin{figure}[ht!]
    \centering
    \includegraphics[width=\linewidth]{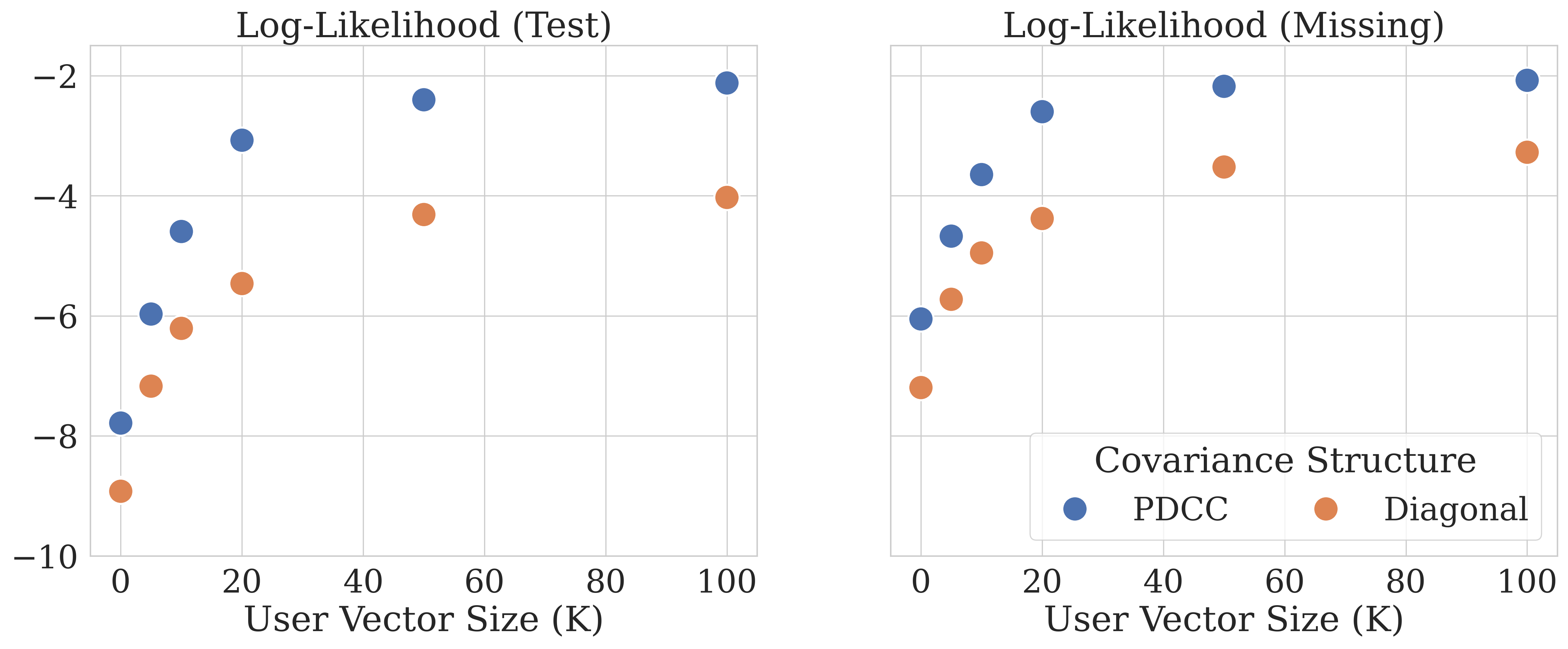}
    \caption{The effect of user vector size $K$ on the modelling performance. $K$=0 means no conditioning on users.}
    \label{fig:Number of LDA Topics}
\end{figure}

\begin{figure}[ht!]
    \centering
    \includegraphics[width=\linewidth]{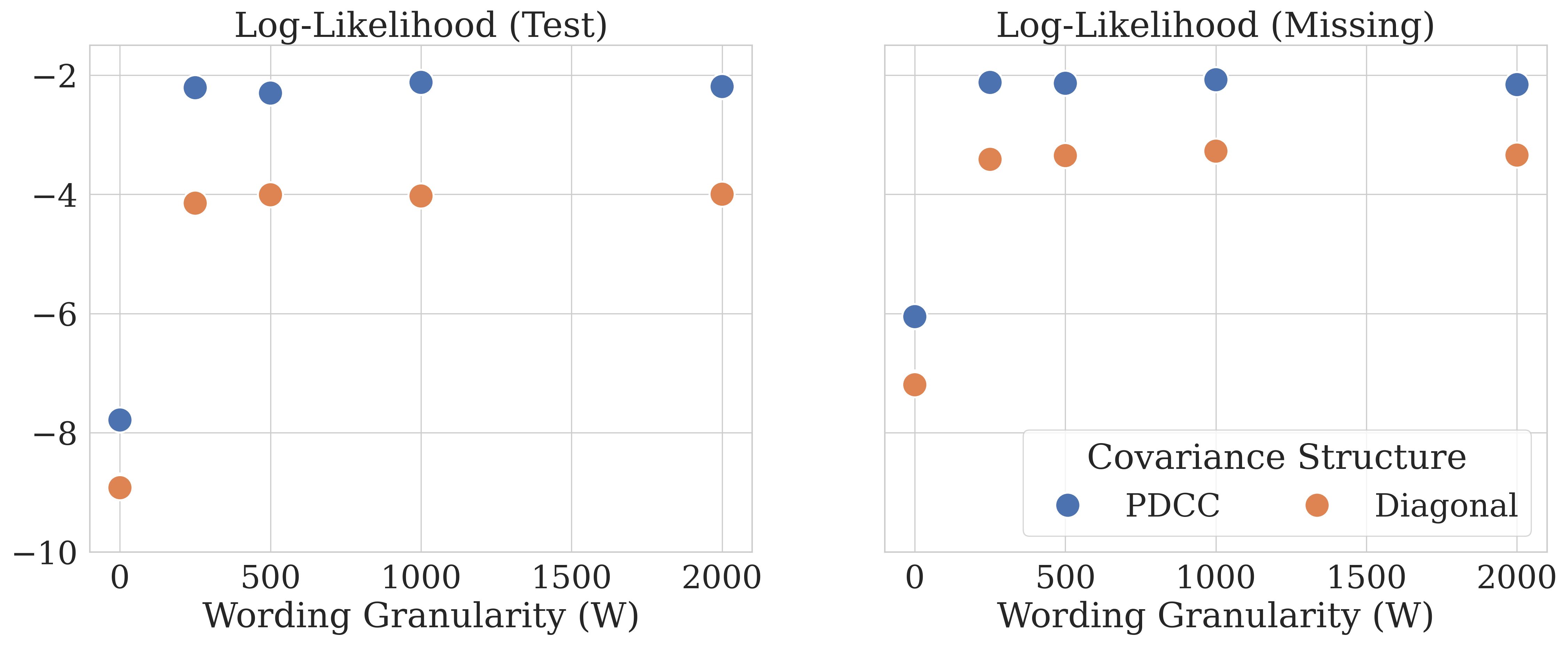}
    \caption{The effect of wording granularity $W$ on the modelling performance. $W$=0 means no conditioning on users.}
    \label{fig:Number of LDA Clusters}
\end{figure}

In order to analyse the effect of user embeddings, two independent hyper-parameter sweeps were performed: 
\begin{itemize}
    \item $K \in \{0,~5,~10,~20,~50,~100\}$
    \item $W \in \{0,~250,~500,~1000,~2000\}$
\end{itemize}
The average log-likelihood scores for the user vector size $K$ and wording granularity $W$ sweeps are given in Fig. \ref{fig:Number of LDA Topics} and \ref{fig:Number of LDA Clusters}, respectively.

The first and most obvious conclusion from Fig. \ref{fig:Number of LDA Topics} is that the introduction of user embeddings boosts the performance considerably. This highlights the efficacy of utilising user information in modelling multi-user datasets. Furthermore, it can be seen that the performance improvement is nearly monotonic with the increasing $K$, until it saturates. Practitioners can select among these values to avoid overwhelming neural network input sizes with unnecessarily large user vectors. 

On the other hand, Fig. \ref{fig:Number of LDA Clusters} reveals that cluster sizes above 250 are sufficiently effective on the performance of this dataset. Since this is a computationally light step in the user embedding scheme and does not affect the neural networks' input sizes, moderately high granularities can be used for wording.

Another result presented in these figures is the performance improvement associated with PDCC. As shown, using a pattern dictionary with $V=100$ consistently improves performance compared to a diagonal covariance structure. This demonstrates that GUIDE-VAE achieves significant performance improvements by leveraging both the user embeddings and PDCC.

\begin{figure}[t!]
    \centering
    \includegraphics[width=\linewidth]{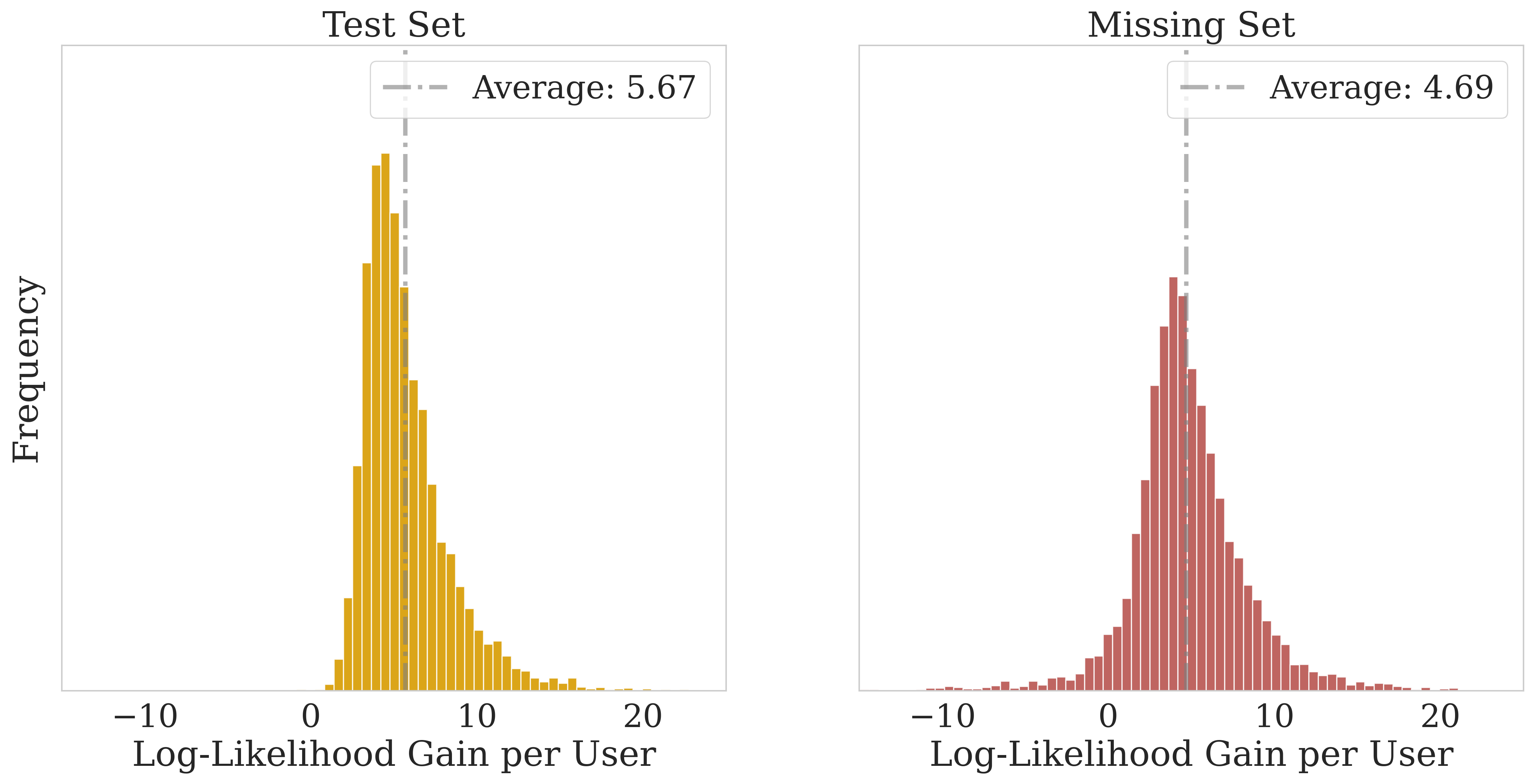}
    \caption{Distributions of log-likelihood gains come with conditioning on user embeddings, per user.}
    \label{fig:likelihood_gain_histogram}
\end{figure}

The performance improvement from user embeddings holds not only globally but also per user. We demonstrate this by calculating the improvement log-likelihood scores individually for each user. We conducted this experiment with a user-guided ($K$=100) and an unguided model by calculating the log-likelihood scores of individual testing ($\mathbb{X}^\text{test}_u = \mathbb{X}^\text{test} \cap \mathbb{X}_u$) and missing datasets ($\mathbb{X}^\text{missing}_u = \mathbb{X}^\text{missing} \cap \mathbb{X}_u$). Then, we subtracted the individual unguided scores from the guided scores to calculate the log-likelihood gain per user. Histograms of these gains for both sets are given in Fig. \ref{fig:likelihood_gain_histogram}. We can see that introducing user embeddings significantly improves the modelling of some users more than others, but user-to-user generalisation is positive for all users in the test set. For comparison, the negative values in the log-likelihood gain on the missing set show that out-of-distribution generalisation is (expectedly) not guaranteed with GUIDE-VAE.

\subsection{Effect of PDCC} 
To further investigate the effect of the pattern dictionary parameters, further hyper-parameter scans were carried out for all possible combinations of the pattern dictionary size $V$ and the lower-bound $\varepsilon$ for the sets
\begin{itemize}
    \item $V \in \{0,~25,~50,~75,~100,~125,~150 \}$,
    \item $\varepsilon \in \{10^{-5},~10^{-4},~10^{-3}\}$.
\end{itemize}
As before, a value of 0 means no employment of PDCC, i.e. a diagonal covariance matrix structure. The value of $\varepsilon$ is not relevant for this case.

\begin{figure}[t!]
    \centering
    \includegraphics[width=\linewidth]{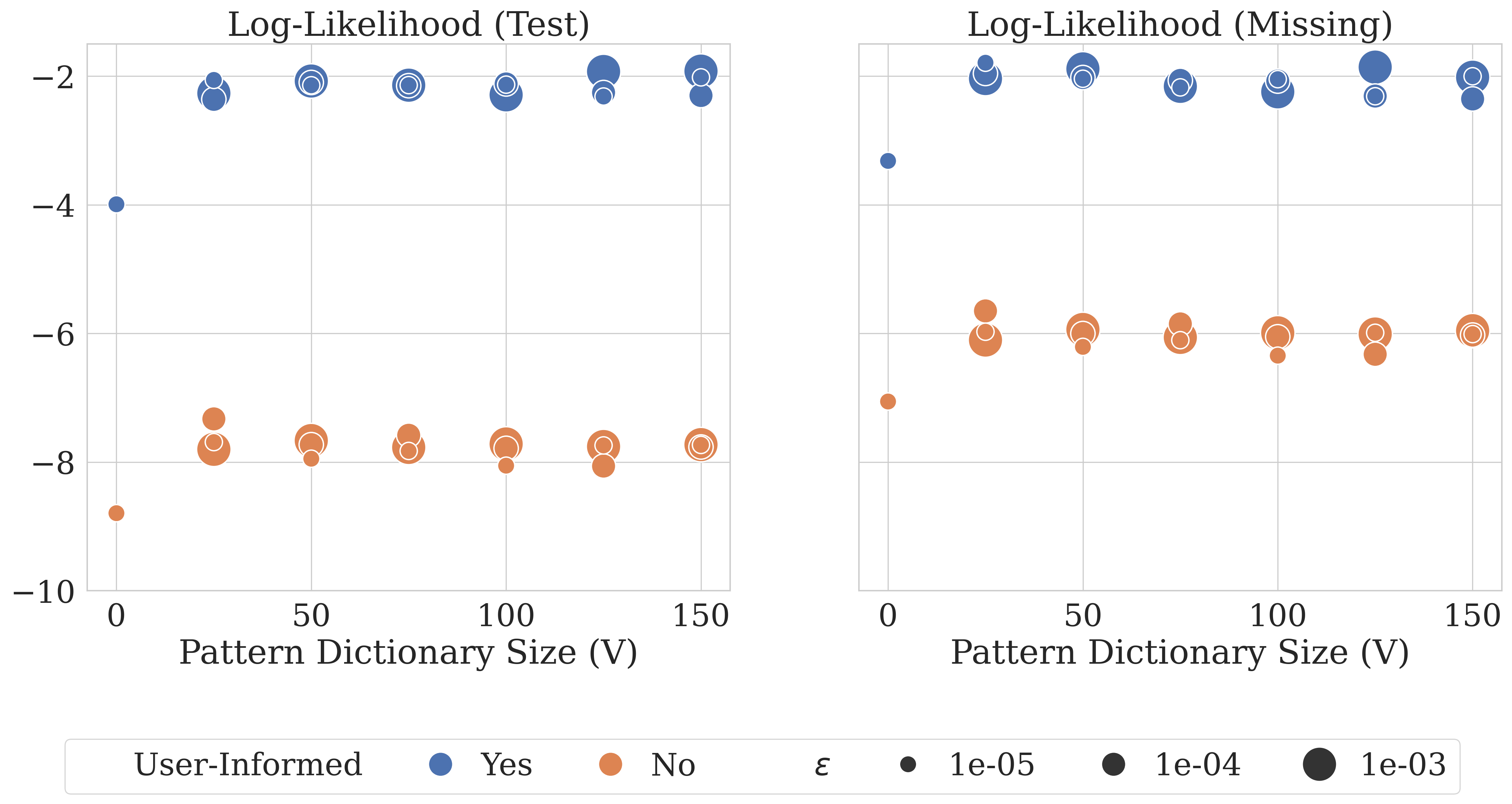}
    \caption{The effect of pattern dictionary size $V$ on the modelling performance. $V$=0 corresponds to the diagonal covariance matrix.}
    \label{fig:Pattern Dictionary Size}
\end{figure}

The results of this sweep are given in Fig. \ref{fig:Pattern Dictionary Size}. In line with the previous result, the performance improvement associated with PDCC is evident. On the other hand, we observe that the selection of neither the dictionary size nor the lower bound results in a significant change in performance.

\begin{figure}[t!]
    \centering
    \includegraphics[width=\linewidth]{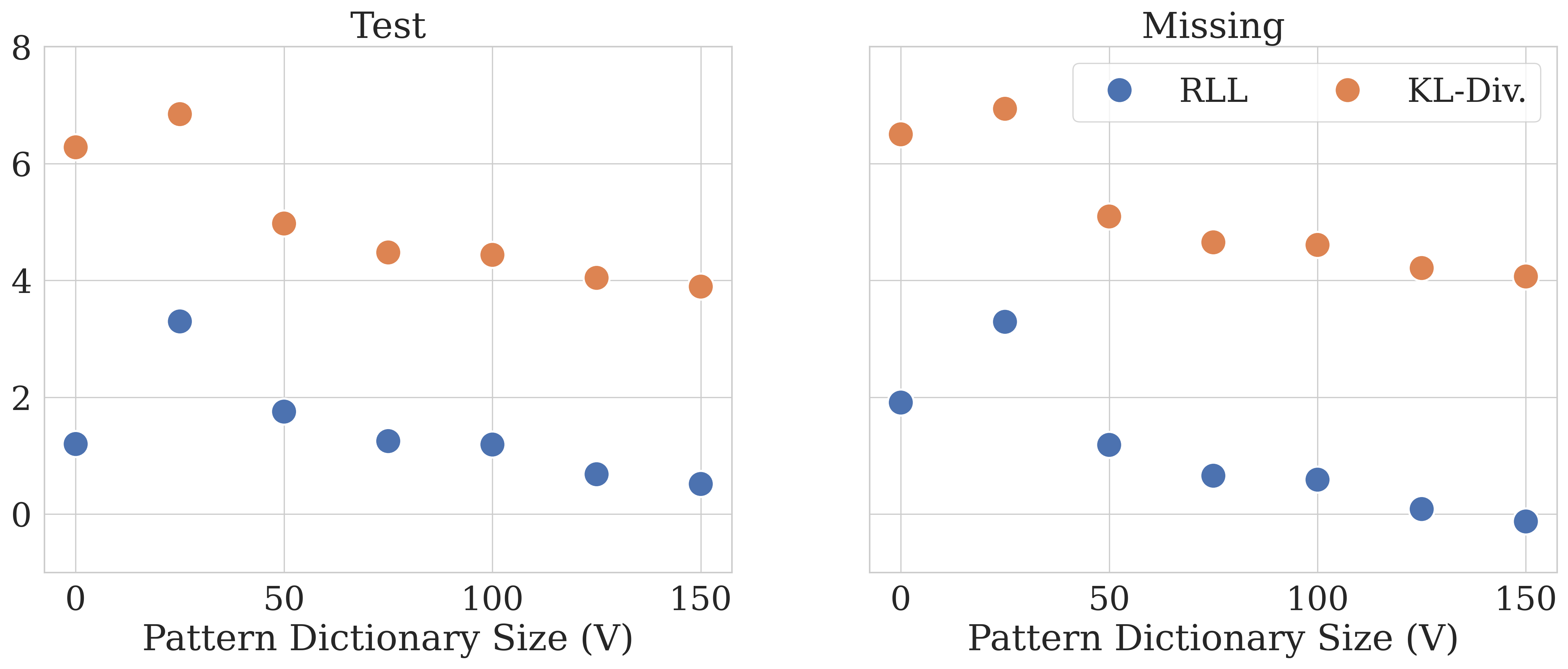}
    \caption{The effect of pattern dictionary size $V$ on reconstruction log-likelihood (RLL) and Kullback-Leibler divergence (KL-Div.). $V$=0 corresponds to the diagonal covariance matrix.}
    \label{fig:dict_size_vs_rll-kl}
\end{figure}

A natural question is why a larger dictionary size does not significantly affect performance, despite offering higher representational power, as we claimed earlier. To answer this, we must investigate the two components of ELBO given in \eqref{eq:elbo}, namely the reconstruction log-likelihood and Kullback-Leibler divergence. These two metrics were calculated for trained models with varying $V$, and the results are given in Fig. \ref{fig:dict_size_vs_rll-kl}. As shown, increasing the dictionary size decreases both the reconstruction log-likelihood and the Kullback-Leibler divergence. It indicates that the larger pattern dictionary gradually takes over the burden of representation from the latent space. This can also be observed from the decreasing deviation on the samples with the pattern dictionary size in Fig. \ref{fig:sample_comparison}. In other words, the model gets more confident about the likelihood distribution imposed by the conditions, and the latent variable $\mathbf{z}$ poses a minimal effect on generation. In other words, a larger pattern dictionary results in a more consistent sampling process.

\subsection{Effect of missingness}
Lastly, we investigate the effect of the missing set size on the performance by sweeping the data availability $b$ as
\begin{itemize}
    \item $b = \{2,~3,~5,~10,~20,~30,~50\}$
\end{itemize}
omitting an average of $\mathbb{E}\left[M_u\right]=365\frac{0.85}{0.85+b}$ per user. Since this amputation process is conducted at the very beginning of the entire training pipeline and is inherently stochastic, relying on a single amputation scenario can be fallacious. Thus, this process is repeated three times with different random number generator seeds for each hyper-parameter configuration.

\begin{figure}[t!]
    \centering
    \includegraphics[width=\linewidth]{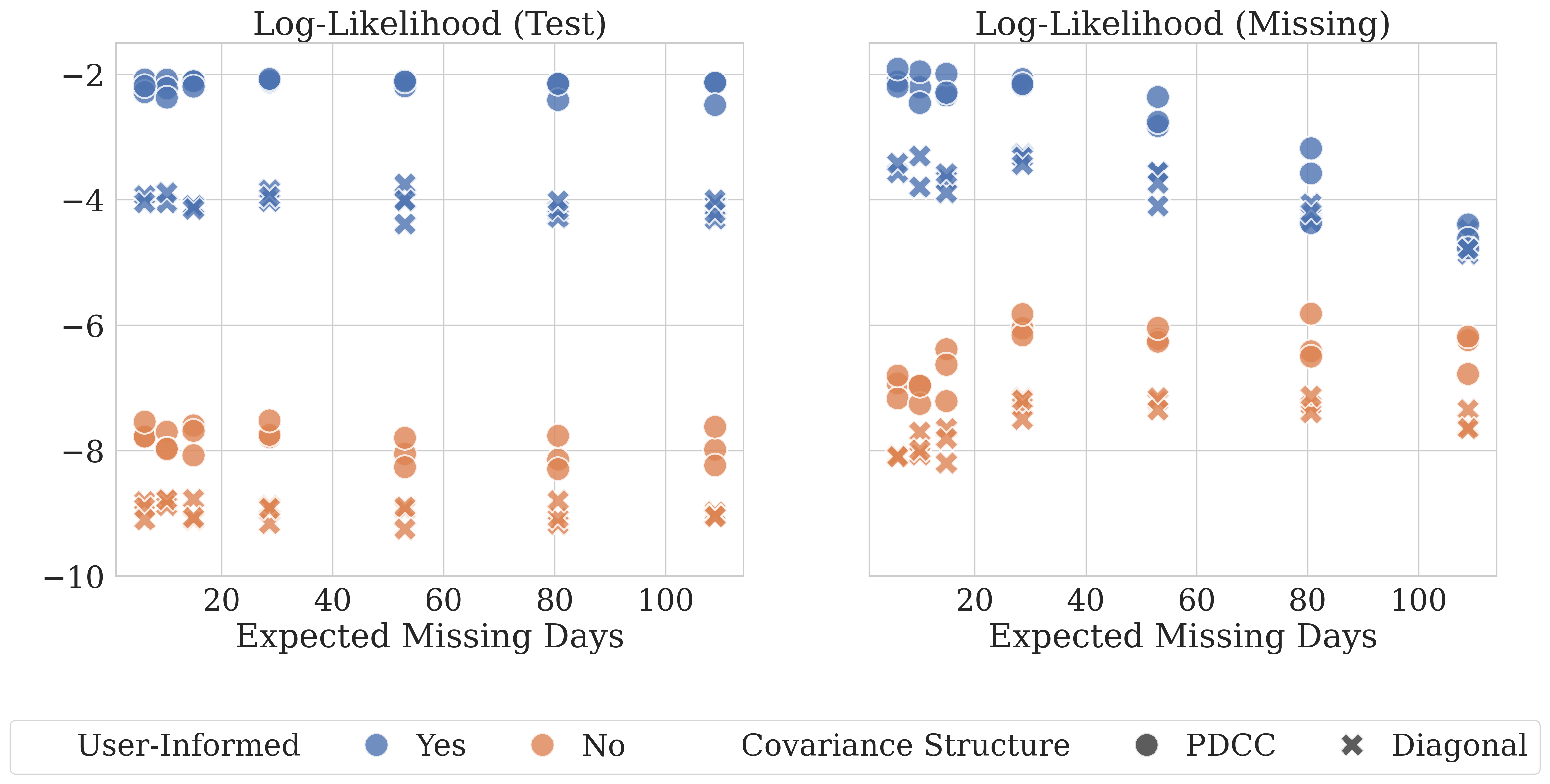}
    \caption{The effect of expected missing days (controlled by $b$) on the modelling performance.}
    \label{fig:Expected Missing Days}
\end{figure}

Fig. \ref{fig:Expected Missing Days} summarises the model performance as a function of the number of missing days. One can immediately spot the near-monotonic decline in performance with expected missing days on the missing set when the user embeddings are provided, while still being superior to its unguided counterparts. On the other hand, performance on the test set appears robust to the size of the training data. This is expected since the testing set is a fixed portion (20\%) of the non-missing part, and the training set is large enough  ($\sim$1M profiles, at worst) to cover the data space of all users.

The robust behaviour in testing performance suggests that GUIDE-VAE can be used confidently for synthetic data generation tasks even under severe missingness. Unfortunately, this is not the case for imputation performance, as the missingness is biased toward the first days of the year. The increasing number of missing days per user results in a lack of information about these periods, and the improved modelling capacity that comes with user embeddings inhibits the out-of-distribution generalisation power of GUIDE-VAE. Thus, we advise practitioners to use GUIDE-VAE with caution for extrapolative missing value imputation tasks.

\section{Conclusion}

In this work, we introduced GUIDE-VAE, a novel framework for generating realistic, user-guided time-series data in multi-user settings. First, we proposed an LDA-based probabilistic embedding scheme for user-specific conditioning, which can be applied to any conditional generative model. This approach effectively addresses the challenges of user-agnostic data generation and data imbalance by incorporating user identities, leading to significant performance improvements. Second, we introduced PDCC to enhance the realism of generated data, particularly by VAEs. PDCC mitigates the common issue of white noise in outputs, enabling the model to capture complex feature dependencies and generate realistic time-series data.

We combined these contributions in the GUIDE-VAE model and demonstrated its effectiveness on a multi-user smart meter dataset. We considered two tasks: synthetic data generation and missing records imputation. The results demonstrated significant improvements over conventional CVAE models, both in terms of modelling performance and sample quality. Overall, GUIDE-VAE offers a powerful solution for generating realistic and user-guided data.

In future work, we plan to showcase the effect of user embeddings in other conditional generative models. Additionally, we aim to explore other applications, such as forecasting, and other multi-user datasets, including those in healthcare and finance.

\bibliographystyle{IEEEtran}
\bibliography{references.bib}

\begin{thebibliography}{10}
\providecommand{\url}[1]{#1}
\csname url@samestyle\endcsname
\providecommand{\newblock}{\relax}
\providecommand{\bibinfo}[2]{#2}
\providecommand{\BIBentrySTDinterwordspacing}{\spaceskip=0pt\relax}
\providecommand{\BIBentryALTinterwordstretchfactor}{4}
\providecommand{\BIBentryALTinterwordspacing}{\spaceskip=\fontdimen2\font plus
\BIBentryALTinterwordstretchfactor\fontdimen3\font minus \fontdimen4\font\relax}
\providecommand{\BIBforeignlanguage}[2]{{%
\expandafter\ifx\csname l@#1\endcsname\relax
\typeout{** WARNING: IEEEtran.bst: No hyphenation pattern has been}%
\typeout{** loaded for the language `#1'. Using the pattern for}%
\typeout{** the default language instead.}%
\else
\language=\csname l@#1\endcsname
\fi
#2}}
\providecommand{\BIBdecl}{\relax}
\BIBdecl

\bibitem{qian2023synthetic}
Z.~Qian, T.~Callender, B.~Cebere, S.~M. Janes, N.~Navani, and M.~van~der Schaar, ``Synthetic data for privacy-preserving clinical risk prediction,'' \emph{medRxiv}, pp. 2023--05, 2023.

\bibitem{peng2022pocket2mol}
X.~Peng, S.~Luo, J.~Guan, Q.~Xie, J.~Peng, and J.~Ma, ``Pocket2mol: Efficient molecular sampling based on 3d protein pockets,'' in \emph{International Conference on Machine Learning}.\hskip 1em plus 0.5em minus 0.4em\relax PMLR, 2022, pp. 17\,644--17\,655.

\bibitem{yoon2020anonymization}
J.~Yoon, L.~N. Drumright, and M.~Van Der~Schaar, ``Anonymization through data synthesis using generative adversarial networks (ads-gan),'' \emph{IEEE journal of biomedical and health informatics}, vol.~24, no.~8, pp. 2378--2388, 2020.

\bibitem{kingma2013auto}
D.~P. Kingma, ``Auto-encoding variational bayes,'' \emph{arXiv preprint arXiv:1312.6114}, 2013.

\bibitem{sohn2015learning}
K.~Sohn, H.~Lee, and X.~Yan, ``Learning structured output representation using deep conditional generative models,'' \emph{Advances in neural information processing systems}, vol.~28, 2015.

\bibitem{bredell2023explicitly}
G.~Bredell, K.~Flouris, K.~Chaitanya, E.~Erdil, and E.~Konukoglu, ``Explicitly minimizing the blur error of variational autoencoders,'' \emph{arXiv preprint arXiv:2304.05939}, 2023.

\bibitem{chai2024defining}
S.~Chai, G.~Chadney, C.~Avery, P.~Grunewald, P.~Van~Hentenryck, and P.~L. Donti, ``Defining `good': Evaluation framework for synthetic smart meter data,'' \emph{arXiv preprint arXiv:2407.11785}, 2024.

\bibitem{sun2023generating}
C.~Sun, J.~van Soest, and M.~Dumontier, ``Generating synthetic personal health data using conditional generative adversarial networks combining with differential privacy,'' \emph{Journal of Biomedical Informatics}, vol. 143, p. 104404, 2023.

\bibitem{zhang2019deep}
S.~Zhang, L.~Yao, A.~Sun, and Y.~Tay, ``Deep learning based recommender system: A survey and new perspectives,'' \emph{ACM computing surveys (CSUR)}, vol.~52, no.~1, pp. 1--38, 2019.

\bibitem{chen2022constructing}
X.~Chen, C.~Zanocco, J.~Flora, and R.~Rajagopal, ``Constructing dynamic residential energy lifestyles using latent dirichlet allocation,'' \emph{Applied Energy}, vol. 318, p. 119109, 2022.

\bibitem{almansoori2022padpaf}
A.~J. Almansoori, S.~Horváth, and M.~Takáč, ``Padpaf: Partial disentanglement with partially-federated gans,'' 2022.

\bibitem{blei2003latent}
D.~M. Blei, A.~Y. Ng, and M.~I. Jordan, ``Latent dirichlet allocation,'' \emph{Journal of machine Learning research}, vol.~3, no. Jan, pp. 993--1022, 2003.

\bibitem{van2021decaf}
B.~Van~Breugel, T.~Kyono, J.~Berrevoets, and M.~Van~der Schaar, ``Decaf: Generating fair synthetic data using causally-aware generative networks,'' \emph{Advances in Neural Information Processing Systems}, vol.~34, pp. 22\,221--22\,233, 2021.

\bibitem{hans2024tabular}
S.~Hans, A.~Sanghi, and D.~Saha, ``Tabular data synthesis with gans for adaptive ai models,'' in \emph{Proceedings of the 7th Joint International Conference on Data Science \& Management of Data}, 2024, pp. 242--246.

\bibitem{Dorta_2018_CVPR}
G.~Dorta, S.~Vicente, L.~Agapito, N.~D.~F. Campbell, and I.~Simpson, ``Structured uncertainty prediction networks,'' in \emph{Proceedings of the IEEE Conference on Computer Vision and Pattern Recognition (CVPR)}, June 2018.

\bibitem{langley2022structured}
J.~Langley, M.~Monteiro, C.~Jones, N.~Pawlowski, and B.~Glocker, ``Structured uncertainty in the observation space of variational autoencoders,'' \emph{Transactions on Machine Learning Research}, 2022.

\bibitem{kaufman2009finding}
L.~Kaufman and P.~J. Rousseeuw, \emph{Finding groups in data: an introduction to cluster analysis}.\hskip 1em plus 0.5em minus 0.4em\relax John Wiley \& Sons, 2009.

\bibitem{muandet2017kernel}
K.~Muandet, K.~Fukumizu, B.~Sriperumbudur, B.~Sch{\"o}lkopf \emph{et~al.}, ``Kernel mean embedding of distributions: A review and beyond,'' \emph{Foundations and Trends{\textregistered} in Machine Learning}, vol.~10, no. 1-2, pp. 1--141, 2017.

\bibitem{horn2012matrix}
R.~A. Horn and C.~R. Johnson, \emph{Matrix analysis}.\hskip 1em plus 0.5em minus 0.4em\relax Cambridge university press, 2012.

\bibitem{quesada2024electricity}
C.~Quesada, L.~Astigarraga, C.~Merveille, and C.~E. Borges, ``An electricity smart meter dataset of spanish households: insights into consumption patterns,'' \emph{Scientific Data}, vol.~11, no.~1, p.~59, 2024.

\bibitem{wang2022contextual}
C.~Wang, S.~H. Tindemans, and P.~Palensky, ``Generating contextual load profiles using a conditional variational autoencoder,'' in \emph{2022 IEEE PES Innovative Smart Grid Technologies Conference Europe (ISGT-Europe)}, 2022, pp. 1--6.

\bibitem{kingma2014adam}
D.~P. Kingma, ``Adam: A method for stochastic optimization,'' \emph{arXiv preprint arXiv:1412.6980}, 2014.

\end{thebibliography}

\end{document}